\documentclass[11pt]{article}

\usepackage[preprint]{acl}
\usepackage{xcolor}

\usepackage{times}
\usepackage{latexsym}
\usepackage[T1]{fontenc}
\usepackage[utf8]{inputenc}
\usepackage{microtype}
\usepackage{inconsolata}
\usepackage{subcaption}
\usepackage{graphicx}
\usepackage{enumitem}
\usepackage{xspace}
\newcommand{\method}{{\textsc{HIPPO}}\xspace}
\usepackage{amsmath}
\usepackage{bm}

\definecolor{amber}{HTML}{F39C12}
\newcommand{\amber}[1]{\textcolor{amber}{#1}}
\definecolor{emerald}{HTML}{27AE60}
\newcommand{\myGreen}[1]{\textcolor{emerald}{\textbf{#1}}}

\usepackage{amsfonts}
\usepackage{algorithm}
\usepackage{algorithmic}
\usepackage{multirow}
\usepackage{booktabs}
\usepackage{amsthm}

\usepackage[table]{xcolor}
\definecolor{desc}{RGB}{99,178,238}
\definecolor{acc}{RGB}{118,218,145}
\definecolor{rej}{RGB}{248,149,136}
\definecolor{uclablue}{rgb}{0.15,0.45,0.68}
\definecolor{c0}{RGB}{0,120,212}
\definecolor{myblue}{RGB}{39, 116, 174}
\newcommand{\blue}{\cellcolor{uclablue!15}}
\title{
    \raisebox{-0.2\height}{\includegraphics[width=0.1\textwidth]{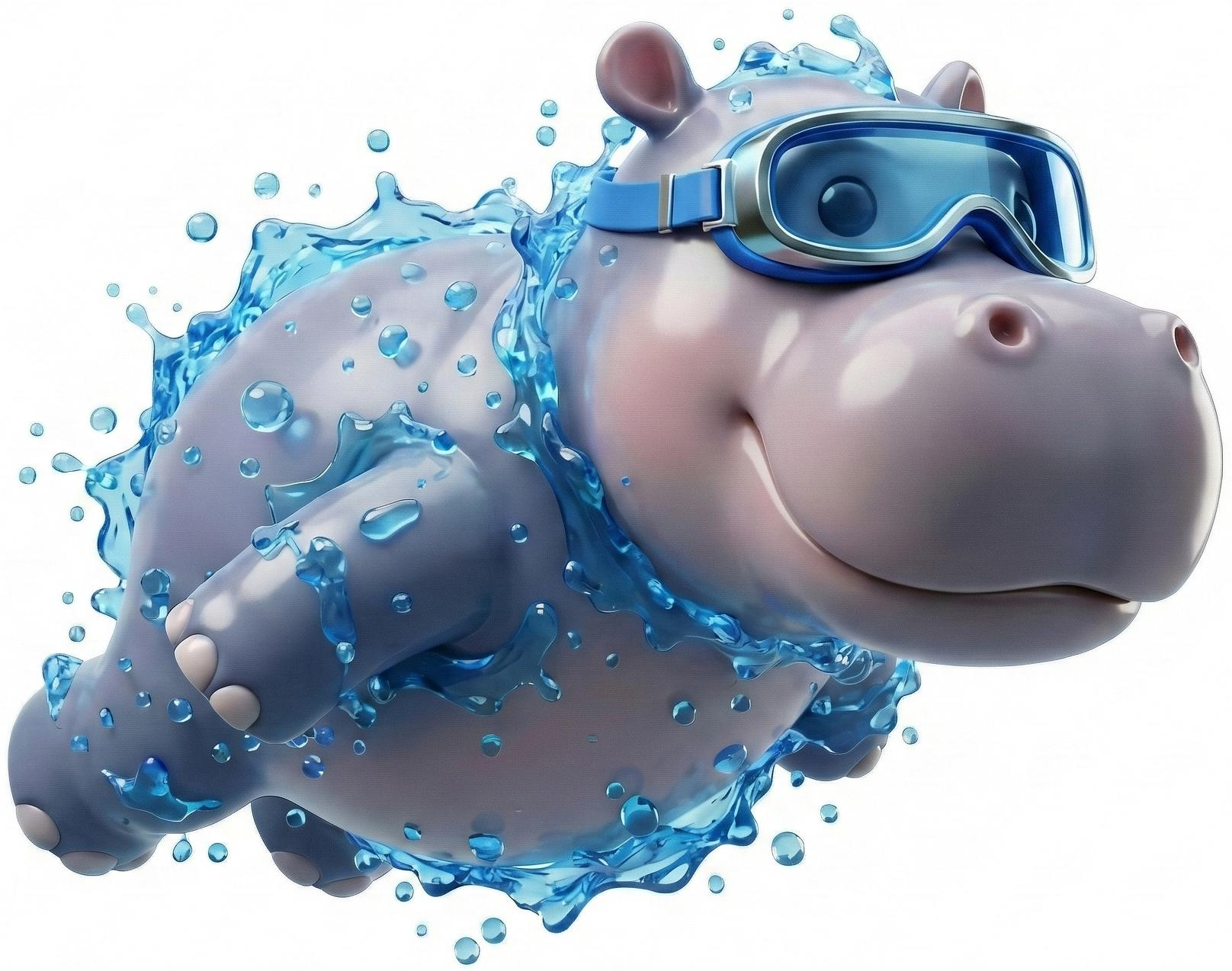}} 
    \method: Accelerating Video Large Language Models Inference via Holistic-aware Parallel Speculative Decoding
}

\author{
    Qitan Lv$^{1,2}$\thanks{Equal contribution.} \quad
    Tianyu Liu$^{1,2}$\footnotemark[1] \quad
    Wen Wu$^{2}$ \quad
    Xuenan Xu$^{2}$ \quad
    Bowen Zhou$^{2,3}$ \quad
    Feng Wu$^{1}$ \quad
    Chao Zhang$^{2,3}$\thanks{The corresponding author.} \\
    $^1$University of Science and Technology of China \quad
    $^2$Shanghai AI Laboratory \\
    $^3$Department of Electronic Engineering, Tsinghua University, China \\
    \texttt{\{qitanlv, tianyu\_liu\}@mail.ustc.edu.cn} \quad
    \texttt{fengwu@ustc.edu.cn} \\
    \texttt{\{wuwen, xuxuenan, zbw, zhangchao\}@pjlab.org.cn}
}

\begin{document}
\maketitle

\begin{abstract}

Speculative decoding (SD) has emerged as a promising approach to accelerate LLM inference without sacrificing output quality. Existing SD methods tailored for video-LLMs primarily focus on pruning redundant visual tokens to mitigate the computational burden of massive visual inputs. 
However, existing methods do not achieve inference acceleration comparable to text-only LLMs.
We observe from extensive experiments that this phenomenon mainly stems from two limitations: {(i)} their pruning strategies inadequately preserve visual semantic  tokens, degrading draft quality and acceptance rates; {(ii)} even with aggressive pruning (e.g., 90\% visual tokens removed), the draft model's remaining inference cost limits overall speedup. 
To address these limitations, we propose \method, a general \textbf{h}olist\textbf{i}c-aware \textbf{p}arallel s\textbf{p}eculative dec\textbf{o}ding framework.
Specifically, \method proposes (i) a semantic-aware token preservation method, which fuses global attention scores with local visual semantics to retain semantic information at high pruning ratios; (ii) a video parallel SD algorithm that decouples and overlaps draft generation and target verification phases.
Experiments on four video-LLMs across six benchmarks demonstrate \method's effectiveness, yielding up to $3.51\times$ speedup compared to vanilla auto-regressive decoding.

\end{abstract}

\section{Introduction}


Recent advances in video Large Language Models (video-LLMs)~\citep{videosalmonn2, qwen25vl, qwen3vl} extend large language models to the video domain, achieving strong performance on video question answering~\citep{videomme} and captioning~\citep{vdc} tasks. However, the auto-regressive, token-by-token generation process imposes significant inference latency. This challenge is exacerbated in the video domain by the sheer volume of input tokens, which
  leads to substantial computational and memory overhead during attention calculation~\citep{specvlm}.


\begin{figure*}[t]
    \centering
    \includegraphics[width=1.8\columnwidth]{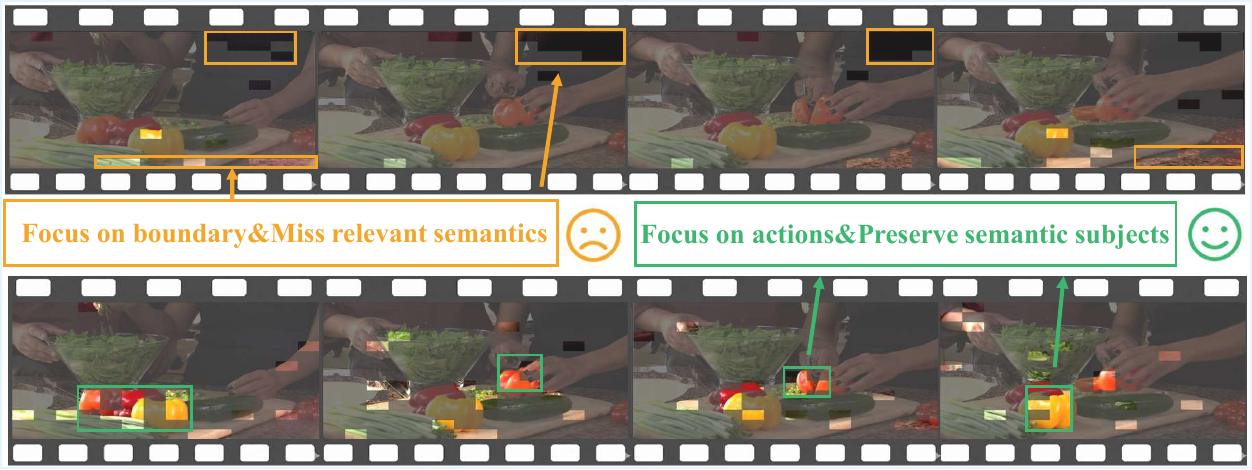}
    \caption{Visualization of video frames under 90\% pruning ratio. (i) The upper part attention-only pruning suffers from the position bias phenomenon, where inflated attention weights are allocated to non-informative boundary tokens \textbf{\amber{(highlighted by orange boxes)}}. (ii) The lower part semantic-aware token preservation preserves informative visual tokens \textbf{\myGreen{(highlighted by green boxes)}}, retaining the core action despite the high pruning ratio.}
    \label{figs:fig1}
\end{figure*}



To this end, extensive research has explored token pruning strategies to accelerate video-LLM inference. These methods exploit spatiotemporal redundancy in video tokens by pruning less important tokens based on attention scores, thereby reducing computational and memory costs during inference~\citep{prunevid, holov, specvlm}. However, by removing tokens from the input, they inevitably incur information loss. This is particularly critical in high-resolution, high-frame-rate long video understanding tasks, where the disruption of spatiotemporal continuity can lead to biased video comprehension~\citep{videomme}.

To mitigate this information loss, \textit{Speculative Decoding} (SD)~\citep{sd1} has emerged as a lossless acceleration approach, where a lightweight draft model proposes multiple candidate tokens that are then verified in parallel by the target model. To adapt SD for the video domain, existing methods mainly prune the draft model's visual tokens using attention scores from the target model.  These methods preserve lossless decoding, as the target model receives the full input for verification~\citep{specvlm, std}.

Despite their benefits, existing video SD methods suffer from significant limitations that prevent them from being as effective as their text-only counterparts. We identify two key limitations: \textbf{(i)} attention-based token selection suffers from position bias~\citep{attensink, attensink2}, where frame boundary tokens receive inflated attention weights regardless of their semantic relevance.  As shown in Figure~\ref{figs:fig1} (top), under 90\% pruning, this leads to retaining non-informative boundary tokens (please refer to Section~\ref{sec:motivation} for statistical results).  \textbf{(ii)} even with aggressive pruning (e.g., 90\%), the draft model still incurs substantial computational overhead. For instance, LLaVA-OneVision~\citep{llavaov} encodes each frame into $196$ tokens, yielding over 1M tokens for a two-minute video at $60$ FPS. After 90\% pruning, the draft model still processes over 100K tokens, resulting in non-negligible inference latency that limits the speedup of serial SD.

  To address these limitations, we propose a general framework, namely \textbf{h}olist\textbf{i}c-aware \textbf{p}arallel s\textbf{p}eculative dec\textbf{o}ding  (\method). Specifically, \method addresses the two limitations and proposes two key designs: \textbf{(i)} a semantic-aware token preservation method that integrates \textbf{global attention scores} with \textbf{local visual semantics} to adaptively allocate the pruning budget across spatial crops. As shown in the lower part of Figure~\ref{figs:fig1}, this method preserves semantically informative visual tokens even at high pruning ratios (e.g., 90\%) by prioritizing tokens that are salient under both global and local criteria. This global-local coordination prevents semantic information loss and ensures that the retained tokens encode coherent scene-level context rather than merely isolated salient attention regions. \textbf{(ii)} a video parallel SD algorithm that enables synchronous execution of the draft and target models, effectively overlapping the draft model's inference cost during the verification phase. While the target model verifies candidate tokens from the previous round, the draft model concurrently generates the next round of candidate tokens. This parallel execution allows both models' computations to overlap, thereby mitigating the limited speedup issue that arises when the draft process accounts for a significant portion of the total latency.

We summarize our contributions as follows:

\begin{enumerate}
    
    \item[(i)]  \textbf{Semantic-aware video token preservation.}  \method integrates global attention scores with local visual semantics to adaptively allocate pruning budgets, enabling effective preservation of semantically informative tokens at high pruning ratios while mitigating position bias.

    \item[(ii)] \textbf{Video parallel speculative decoding.} \method proposes a parallel execution framework where the draft model generates candidates concurrently while the target model performs verification, effectively overlapping their computation to hide the draft overhead.

    \item[(iii)]  \textbf{Significant and versatile speedups.} Experiments on four video-LLMs across six benchmarks show that \method achieves up to $3.51\times$ speedup over auto-regressive decoding, demonstrating its effectiveness and versatility.

\end{enumerate}


\section{Related Work}
\label{sec:related}

\subsection{Speculative Decoding for LLMs}

Speculative decoding has emerged as an effective approach to accelerate LLM inference without compromising generation quality. Existing works focus on improving draft model efficiency and increasing acceptance rates~\citep{specinfer, spectr, medusa}. 
 Medusa~\citep{medusa} adds extra decoding heads at the top of the target model to generate drafts. Lookahead~\citep{lookahead} caches the generation trajectory (n-grams) as the drafts. PEARL~\citep{pearl} parallelizes the execution of the draft and target models to achieve superior acceleration. Eagle~\citep{eagle3} balances efficiency and acceptance rates by training a single-layer transformer as the draft model.  
 
 In the video domain, existing methods focus on exploiting the redundancy of visual tokens. They use signals from the target model to prune the draft model's visual input, thereby improving drafting efficiency~\citep{specvlm, std}. Sparse-to-Dense~\citep{std} uses a sparse module to generate tokens with top-K attention in the draft phase, which are then verified in parallel using full attention. SpecVLM~\citep{specvlm} selects tokens guided by target model attention and applies uniform pruning across all frames. However, existing video SD methods suffer from position bias in the target model's attention, limiting their effectiveness. \method addresses this limitation through semantic-aware token preservation.  We also discuss video-LLMs and long-context SD in the text domain, explaining why the latter are not directly applicable to video-LLMs in Appendix~\ref{app:related}.

\subsection{Visual Token Reductions}

Visual inputs exhibit significant spatiotemporal redundancy, incurring high computational and memory costs~\citep{holov}. To mitigate this, extensive research focuses on token compression and pruning in vision transformers~\citep{vit}. In particular, video token reduction has received increasing attention due to the massive volume of tokens generated by video inputs. For instance, FastVID~\citep{fastvid} partitions frames into segments and applies density-based token pruning within each segment. DyCoke~\citep{dycoke} performs token merging across frames and reduces KV cache dynamically. Unlike single-model pruning approaches, \method leverages signals from the target model to guide pruning for the draft model. Therefore, \method is orthogonal to single-model pruning methods and can be combined with them for specific downstream tasks.

\section{Preliminaries}
\label{sec:pre}

\subsection{Notations}


Let $F$ denote the number of video frames. Each token $\mathbf{v}_j \in \mathbb{R}^{d}$ is a visual embedding of dimension $d$. We use $\mathcal{M}_t$ and $\mathcal{M}_d$ to denote the target and draft models, respectively. During speculative decoding, the draft model generates $\gamma$ candidate tokens with logits $q_i$, which the target model verifies using logits $p_i$. We define three scoring functions: $s_{\text{attn}}$ for global attention-based relevance, $s_{\text{temp}}$ for inter-frame temporal redundancy, and $s_{\text{spa}}$ for intra-frame spatial redundancy, which are aggregated to produce the final score $s(\mathbf{v}_j)$.

\subsection{Speculative Decoding}

Let $\bm{x}$ denote an input sequence (prefix). A speculative decoding (SD) step consists of a drafting phase followed by a verification phase. In the drafting phase, the draft model $\mathcal{M}_d$ is invoked $\gamma$ times to generate draft tokens $x_1, x_2, \ldots, x_\gamma$ by forward passes and sampling. We denote the output logit $\mathcal{M}_d\bigl(\bm{x} + [x_1, \ldots, x_{i}]\bigr)$  as $q_{i}$, and sample each draft token as $x_i \sim q_{i}$ for $i = 1, \ldots, \gamma$. In the verification phase, the prefix $\bm{x}$ and the $\gamma$ draft tokens are sent to the target model $\mathcal{M}_t$ for verification. The model $\mathcal{M}_t$ takes $\bm{x} + [x_1, \ldots, x_\gamma]$ as input and outputs logits $p_0, p_1, \ldots, p_\gamma$. SD then sequentially verifies each $x_i$ via speculative sampling~\citep{sd1}, with acceptance probability
\begin{equation}
  \alpha_i =
  \begin{cases}
    1, & \text{if } p_{i}[x_i] \ge q_{i}[x_i],\\[3pt]
    \dfrac{p_{i}[x_i]}{q_{i}[x_i]}, & \text{if } p_{i}[x_i] < q_{i}[x_i].
  \end{cases}
\end{equation}

If SD rejects $x_i$, it resamples a token from
$  \operatorname{norm}\!\bigl(\max(0, p_{i} - q_{i})\bigr);
$
otherwise, SD accepts all draft tokens and samples one additional token from $p_\gamma$. In this way, each SD yields theoretical lossless quality with improved decoding efficiency.

\section{Motivated Experiments}
\label{sec:motivation}

\subsection{Position Bias Issue of Visual Tokens}

\begin{figure}[t]
    \centering
    \includegraphics[width=\columnwidth]{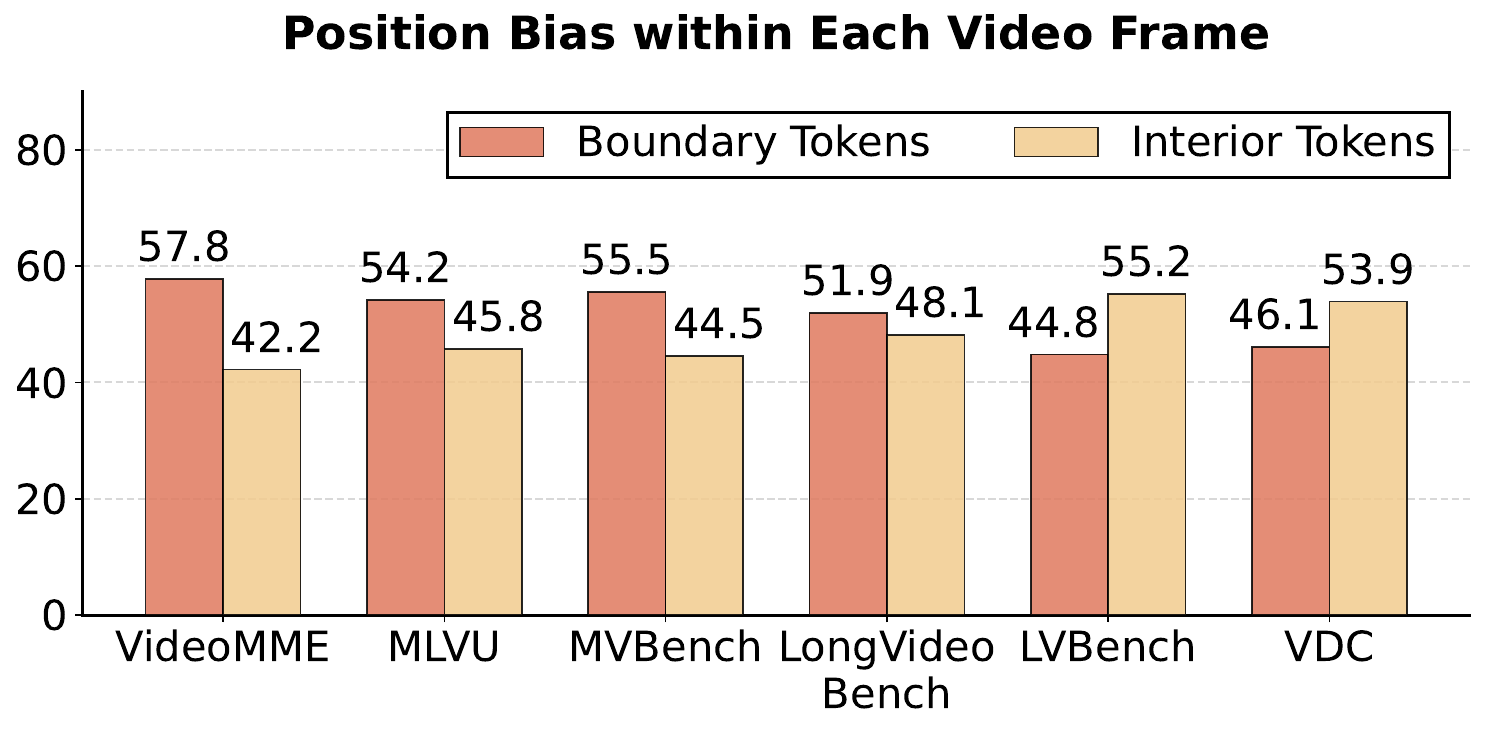}
    \caption{Statistics of per-frame positional bias. }
    \label{figs:position_bias}
\end{figure}

We analyze the attention score distribution of video-SALMONN2+ 72B~\citep{videosalmonn2} on the Video-MME benchmark, revealing a distinct position bias phenomenon. As shown in Figure~\ref{figs:position_bias}, we select the top 10\% video tokens by attention scores and investigate their spatial distribution within individual frames. We define tokens located within a normalized distance of 10\% from the top or bottom edges as boundary tokens (constituting approximately 20\% of the total), while the remaining 80\% are classified as interior tokens. We observe that among the selected tokens in each frame, boundary tokens account for a disproportionate share—exceeding 44\% and peaking at 57.8\%. This suggests that relying solely on attention-based token selection is insufficient for effectively capturing visual semantics.

\begin{figure}[t]
    \centering
    \includegraphics[width=\columnwidth]{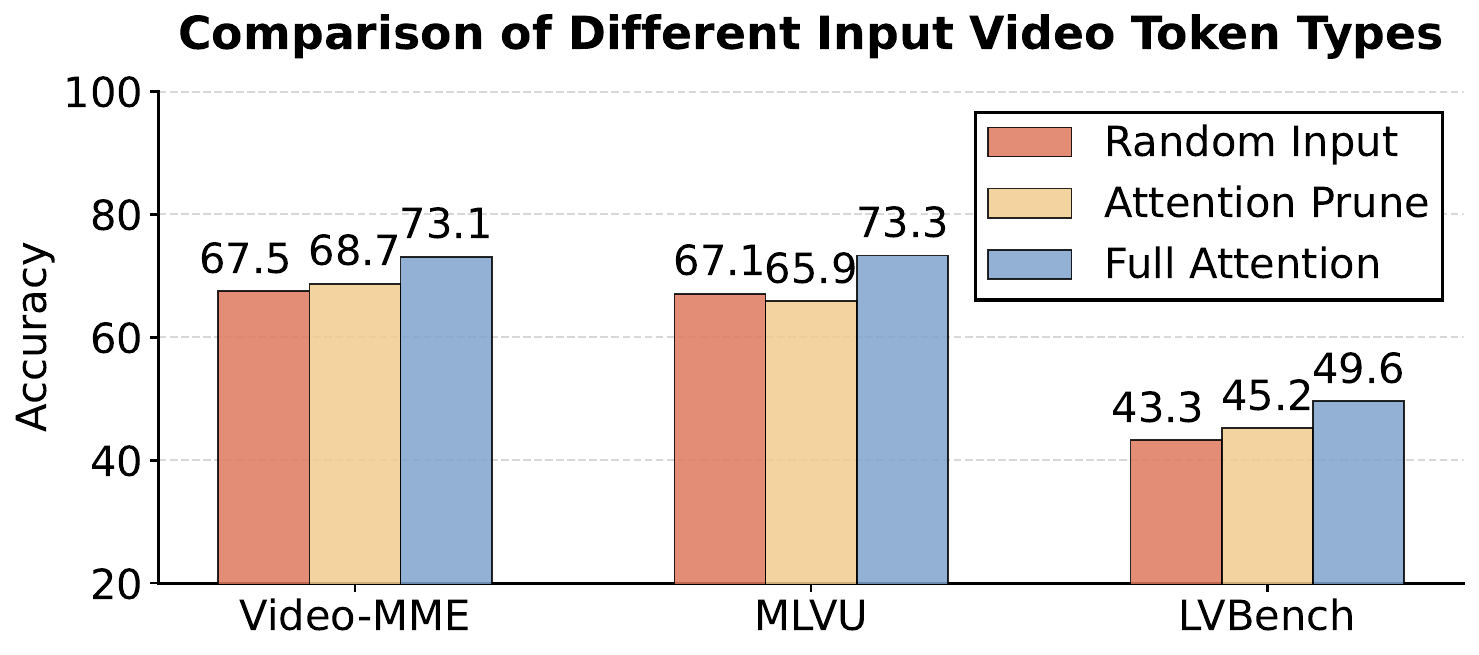}
    \caption{Performance comparison of three visual token inputs: random 10\% sampling, target-model-guided top-10\% attention selection, and full visual tokens.}
    \label{figs:compa}
\end{figure}

   We further conduct a controlled experiment using video-SALMONN2+ 7B as the draft model and video-SALMONN2+ 72B as the target model. We evaluate the draft model's performance under three input configurations: full visual tokens, top-10\% tokens selected via target model attention scores, and a random 10\% selection of visual tokens. We report accuracy on three video-QA benchmarks: Video-MME~\citep{videomme}, MLVU~\citep{mlvu}, and LVBench~\citep{lvbench}. As shown in Figure~\ref{figs:compa}, both the attention-only and random strategies result in performance degradation compared to the full-token baseline. Notably, on MLVU, the random strategy outperforms attention-based pruning. This demonstrates the significance of preserving semantic visual tokens and suggests that attention-only selection can discard semantically important tokens due to position bias.


\begin{figure}[t]
    \centering
    \includegraphics[width=\columnwidth]{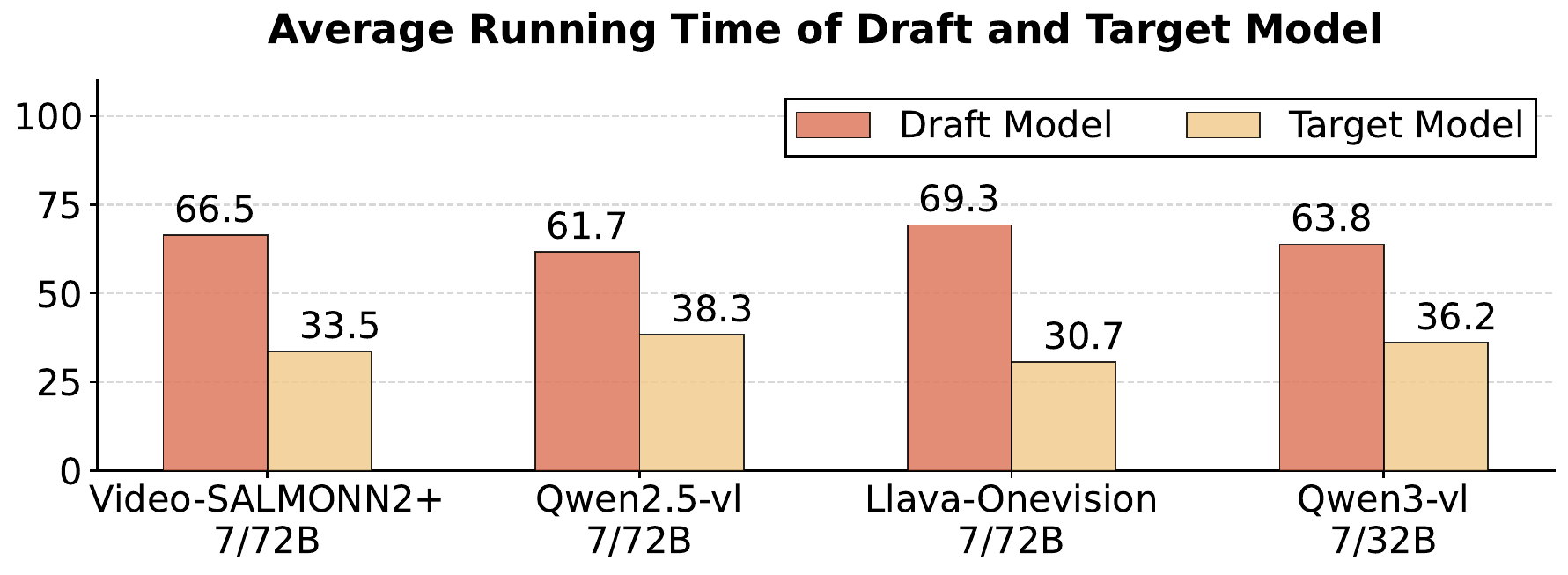}
    \caption{Statistics of average running time ratio of the draft and target model.}
    \label{figs:draft_target}
\end{figure}

\subsection{Non-negligible Draft Latency}

 We conduct a series of real-world experiments to demonstrate that draft model latency is a significant bottleneck in speculative decoding. As shown in Figure~\ref{figs:draft_target}, both the drafting phase and verification phase consume substantial time. At each decoding step, the draft model remains idle during verification while the target model remains idle during drafting. This mutual waiting dilutes the speedup from accepted draft tokens, and the overhead from rejected draft tokens degrades performance.

For example, suppose the target model is three times slower than the draft model, i.e., \(T_{\text{target}} = 3 T_{\text{draft}}\). In each round, the draft model proposes 3 tokens. If all three tokens are accepted, the total computation corresponds to three forward passes of the draft model and one forward pass of the target model to obtain 4 correct tokens (3 draft tokens plus 1 token from the target model's verification). The effective speedup is \(4 T_{\text{target}} / (3 T_{\text{draft}} + 1 T_{\text{target}}) = 2\). However, if the first draft token is rejected, the same amount of computation (three draft forward passes and one target forward pass) yields only 1 correct token (from the target model's verification), and the effective speedup becomes \(1 T_{\text{target}} / (3 T_{\text{draft}} + 1 T_{\text{target}}) = 0.5\), which is significantly worse than vanilla AR decoding. Due to the serial execution of the draft and target models, together with the substantial draft model
  latency under long visual tokens, existing video-SD methods suffer from a fundamental bottleneck.

\section{Method}
\label{sec:method}

\begin{figure*}[t]
    \centering
    \includegraphics[width=\textwidth]{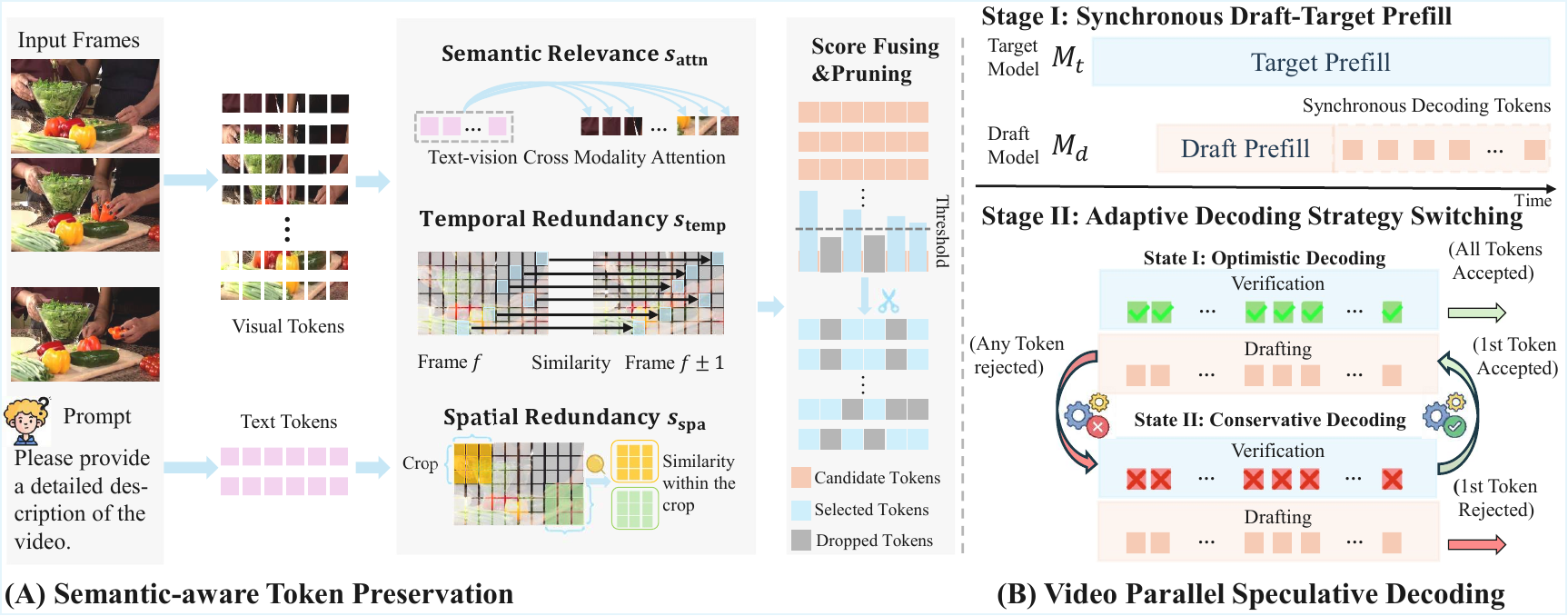}
    \caption{An overview of \method. (A) Semantic-aware Token Preservation: A holistic scoring mechanism integrates global semantic relevance, inter-frame temporal redundancy, and intra-frame spatial redundancy to select a compact subset of informative visual tokens, mitigating position bias. (B) Video Parallel Speculative Decoding: Synchronous Draft-Target Prefill generates initial draft tokens while the target model processes input, followed by an Adaptive Decoding Strategy that dynamically switches between optimistic post-verification and conservative pre-verification modes based on draft quality to maximize parallelism.
}
    \label{fig:model_figure}
\end{figure*}

We propose \method, a \textbf{h}olist\textbf{i}c-aware \textbf{p}arallel s\textbf{p}eculative dec\textbf{o}ding framework that promotes video-LLMs to focus on informative visual content and overlapping the non-negligible draft latency. \method integrates semantic-aware token preservation—which holistically scores tokens across global, temporal, and spatial dimensions to mitigate position bias—and video parallel speculative decoding—which overlaps draft generation with target verification via adaptive strategy switching. An overview of \method is shown in Figure \ref{fig:model_figure}.

\subsection{Semantic-aware Token Preservation}

Consider a video of a news anchor. While the anchor's lip movements, gestures, and expressions exhibit temporal variations, the background and studio layout remain static across adjacent frames. From a spatial perspective, large intra-frame regions—such as the desk and walls—contain little information. This example illustrates the dual nature of video redundancy: \textbf{temporally}, static backgrounds repeat across frames; \textbf{spatially}, uniform regions within each frame carry minimal information. An effective token preservation strategy should capture both dimensions to preserve semantic completeness under aggressive compression.

As demonstrated by motivated experiments in Section~\ref{sec:motivation}, relying solely on attention scores neglects crucial visual information due to position bias~\citep{attensink}. To address this, we propose a semantic-aware scoring mechanism that comprehensively evaluates each video token by integrating three orthogonal signals: \textbf{global semantic relevance}, \textbf{inter-frame temporal redundancy}, and \textbf{intra-frame spatial redundancy}. By scoring and ranking tokens holistically, we identify a compact subset that captures the essential visual semantics for the draft model.

\paragraph{Global Semantic Relevance.}
  To identify tokens aligned with the user prompt, we compute a global semantic relevance score $s_{\text{attn}}$ from language-to-video cross-attention. In our example, if the prompt is ``What is the anchor reporting?'', tokens corresponding to the anchor's face and on-screen graphics should receive high scores. Following prior work~\citep{specvlm}, we leverage KV-based reconstruction to avoid computing the full $O(L^2)$ attention matrix. Specifically, we extract visual keys $\mathbf{K}_{\text{vid}} \in \mathbb{R}^{N_v \times d_k}$ from the prefilled KV cache at minimal cost, obtain text queries $\mathbf{Q}_{\text{txt}} \in \mathbb{R}^{L_{\text{txt}} \times d_k}$, and compute the cross-modal attention matrix $\mathbf{S}_{\text{attn}} = \text{softmax}(\mathbf{Q}_{\text{txt}} \mathbf{K}_{\text{vid}}^T / \sqrt{d_k}) \in \mathbb{R}^{L_{\text{txt}} \times N_v}$. We aggregate weights across layers and heads:
  \begin{equation}
  s_{\text{attn}}(\mathbf{v}_j) = \frac{1}{N_L H L_{\text{txt}}} \sum_{\ell=1}^{N_L} \sum_{h=1}^{H} \sum_{i=1}^{L_{\text{txt}}} [\mathbf{S}_{\text{attn}}^{(\ell,h)}]_{i,j},
  \end{equation}
  where $L_{\text{txt}}$ is the text sequence length, $N_L$ is the number of layers, $H$ is the number of attention heads, $d_k$ is the key dimension, and $\ell, h$ index the layer and head respectively. This score effectively captures prompt-relevant content. However, it suffers from position bias, where boundary tokens receive inflated scores regardless of content.

  \paragraph{Inter-frame Temporal Redundancy.}
  To quantify temporal redundancy, we introduce a temporal score $s_{\text{temp}}$ measuring cross-frame similarity at corresponding spatial locations. For example, tokens at static background positions exhibit high cross-frame similarity, while tokens at the anchor's moving lips show low similarity. For token $\mathbf{v}_j^{f} \in \mathbb{R}^d$ at position $j$ in frame $f$, we compute its cosine similarity with tokens at the same spatial position in adjacent frames:
  \begin{equation}
  s'_{\text{temp}}(\mathbf{v}_j^{f}) = \frac{1}{|\mathcal{N}(f)|} \sum_{f' \in \mathcal{N}(f)} \frac{{\mathbf{v}_j^{f}}^T \mathbf{v}_j^{f'}}{\|\mathbf{v}_j^{f}\| \|\mathbf{v}_j^{f'}\|},
  \end{equation}
  where $\mathcal{N}(f) = \{f-1, f+1\} \cap \{1, \ldots, F\}$ denotes adjacent frames. Since high similarity indicates redundancy (which we want to prune), we invert the score to prioritize temporal variation:
  \begin{equation}
  s_{\text{temp}}(\mathbf{v}_j^{f}) = 1 - s'_{\text{temp}}(\mathbf{v}_j^{f}).
  \end{equation}

\paragraph{Intra-frame Spatial Redundancy.}
  We evaluate the visual complexity of each token within its local crop. For a given crop $c$ containing $M$ tokens, we compute the intra-crop similarity matrix by measuring pairwise relationships among all tokens. Given the $L_2$-normalized embeddings $\mathbf{E}_v^c \in \mathbb{R}^{M \times d}$ in crop $c$, we compute the similarity matrix $\mathbf{S}^c$ as:
  \begin{equation}
  \mathbf{S}^c = \mathbf{E}_v^c ({\mathbf{E}_v^c})^\top.
  \end{equation}
  We capture the spatial complexity of each token $\mathbf{v}_j^{f,c}$ by computing the variance of its similarity distribution within the crop:
  \begin{equation}
  s_{\text{spa}}(\mathbf{v}_j^{f,c}) = \frac{1}{M} \sum_{k=1}^{M} (\mathbf{S}_{j,k}^c - \mu_j^c)^2,
  \end{equation}
  where $\mu_j^c = \frac{1}{M}\sum_{k=1}^{M} \mathbf{S}_{j,k}^c$ is the mean similarity of token $j$ with other tokens in crop $c$, and $M$ is the number of tokens per crop. High variance indicates that the token has diverse connections with others (e.g., anchor's gestures and expressions), suggesting high information content, while low variance suggests uniformity (e.g., desk and walls).

\paragraph{Holistic Score Fusion.}
  The three scores capture complementary aspects: $s_{\text{attn}}$ identifies prompt-relevant semantics but suffers position bias; $s_{\text{temp}}$ detects temporal change but ignores static importance; $s_{\text{spa}}$ measures structural complexity but lacks global semantic context. For each frame $f$, we normalize each score type independently: $\hat{s}(\mathbf{v}_j) = (s(\mathbf{v}_j) - \mu_f) / \sigma_f$, where $\mu_f$ and $\sigma_f$ are the mean and standard deviation of that score type within frame $f$. The final holistic score is:
  \begin{equation}
  s(\mathbf{v}_j) = \hat{s}_{\text{attn}}(\mathbf{v}_j) + \hat{s}_{\text{spa}}(\mathbf{v}_j) + \hat{s}_{\text{temp}}(\mathbf{v}_j).
  \end{equation}
  We finally retain the top-$k$ (by default $10$\%) visual tokens via $s(\mathbf{v}_j)$.


\subsection{Video Parallel Speculative Decoding}
While semantic-aware token preservation effectively reduces the visual input to the draft model, sequential dependencies in standard speculative decoding create mutual waiting: the draft model $\mathcal{M}_d$ must complete all $\gamma$ tokens before the target model $\mathcal{M}_t$ can verify, and $\mathcal{M}_d$ idles during verification while $\mathcal{M}_t$ idles during drafting. Drawing upon parallelization paradigms from text-only LLMs~\citep{pearl}, we propose a \textbf{Video Parallel Speculative Decoding (VPSD)} framework tailored for video-LLMs. As shown in Figure \ref{fig:model_figure}, VPSD transforms the generation pipeline by overlapping draft generation and target verification phases.

  \paragraph{Synchronous Draft-Target Prefill.}
In video-LLMs, the target model's prefill latency is dominated by the computationally expensive vision encoder, significantly exceeding the draft model's prefill time. We exploit this asymmetry during the initial prefill stage: both models process the input through the vision encoder, but $\mathcal{M}_d$ continues auto-regressive decoding until $\mathcal{M}_t$ completes its prefill. For instance, if $\mathcal{M}_t$ requires 0.8s for prefill while $\mathcal{M}_d$ needs only 0.2s, the draft model can generate approximately 12 tokens (assuming 50ms per token) before the target model is ready. It accumulates a larger initial buffer of candidate tokens, effectively hiding the target model's overhead.

  \paragraph{Adaptive Decoding Strategy Switching.}
After the prefilling stage, we introduce an adaptive decoding mechanism that dynamically switches between two strategies based on the verification outcome of the previous step. \textbf{When draft quality is high} (i.e., the previous batch was fully accepted), the system enters an \textbf{optimistic decoding mode}: $\mathcal{M}_d$ speculatively generates the next batch $[x_{\gamma+1}, \ldots, x_{2\gamma}]$ concurrently with $\mathcal{M}_t$'s verification of the current batch $[x_1, \ldots, x_\gamma]$. If all tokens are accepted, the next batch is already available for immediate use; if any rejection occurs, it discards the speculative tokens at no additional cost since the computation was performed in parallel. \textbf{Conversely, when draft quality is low} (i.e., a rejection occurred), it switches to a \textbf{conservative decoding mode}: $\mathcal{M}_t$ verifies the first draft token $x_1$ in parallel with $\mathcal{M}_d$'s generation of the remaining tokens $[x_2, \ldots,
  x_\gamma]$. If $x_1$ is rejected, we immediately abort drafting and discard $[x_2, \ldots, x_\gamma]$, preventing wasted computation; if accepted, the remaining tokens proceed to standard verification.
    This dynamic switching implements an implicit draft length adaptation: aggressive when $\mathcal{M}_d$ produces reliable drafts, cautious when the quality is low. It ensures both models operate in parallel throughout generation, alleviating the mutual waiting issue.

\section{Experiments}
\label{sec:experiments}



\begin{table*}[htbp]
\centering
 \caption{Performance comparison across different backbone models. We report Mean Acceptance Tokens (MAT) and wall-time speedup relative to standard auto-regressive decoding. We \textbf{bold} the best results for each model.}
\label{tab:merged_performance}
\resizebox{\textwidth}{!}{%
\setlength{\tabcolsep}{4pt}
\begin{tabular}{clcccccccccccc}
\toprule
\multirow{2}{*}{\textbf{Backbone Model}} & \multirow{2}{*}{\textbf{Method}} & \multicolumn{2}{c}{\textbf{VideoMME}} & \multicolumn{2}{c}{\textbf{VDC}} & \multicolumn{2}{c}{\textbf{MVBench}} & \multicolumn{2}{c}{\textbf{LongVideoBench}} & \multicolumn{2}{c}{\textbf{MLVU}} & \multicolumn{2}{c}{\textbf{LVbench}} \\
\cmidrule(lr){3-4} \cmidrule(lr){5-6} \cmidrule(lr){7-8} \cmidrule(lr){9-10} \cmidrule(lr){11-12} \cmidrule(lr){13-14}
 & & MAT & Spd. & MAT & Spd. & MAT & Spd. & MAT & Spd. & MAT & Spd. & MAT & Spd. \\
\midrule

\multirow{5}{*}{\shortstack[l]{Video-SALMONN2+\\\qquad 7\&72B}} 
 & Vanilla SD & 2.90 & 1.36$\times$ & 2.76 & 1.41$\times$ & 2.71 & 1.33$\times$ & 2.75 & 1.34$\times$ & 3.30 & 1.53$\times$ & 2.89 & 1.57$\times$\\
 & SD-tree & 3.71 & 1.87$\times$ & 3.64 & 1.73$\times$ & 3.18 & 1.54$\times$ & 3.11 & 1.45$\times$ & 3.86 & 1.83$\times$ & 3.29 & 1.72$\times$ \\
 & SpecVLM & 3.46 & 2.24$\times$ & 3.42 & 2.01$\times$ & 3.00 & 2.09$\times$ & 3.34 & 2.15$\times$ & 4.00 & 2.13$\times$ & 3.42 & 2.28$\times$ \\
 & \blue{\method} & \textbf{\blue{11.82}} & \blue{\textbf{2.85$\times$}} & \textbf{\blue{6.97}} & \blue{\textbf{2.31$\times$}} & \textbf{\blue{12.31}} & \blue{\textbf{2.89$\times$}} & \textbf{\blue{10.06}} & \blue{\textbf{2.74$\times$}} & \textbf{\blue{12.42}} & \blue{\textbf{2.78$\times$}} & \textbf{\blue{12.12}} & \blue{\textbf{2.64$\times$}} \\
\midrule

\multirow{5}{*}{\shortstack[l]{Qwen2.5-VL\\ \quad 7\&72B}} 
  & Vanilla SD & 2.70 & 1.48$\times$ & 3.37 & 1.56$\times$ & 2.69 & 1.61$\times$ & 2.97 & 1.52$\times$ & 3.39 & 1.64$\times$ & 3.11 & 1.57$\times$ \\
 & SD-tree & 3.32 & 1.73$\times$ & 3.64 & 1.72$\times$ & 3.00 & 1.85$\times$ & 3.31 & 1.72$\times$ & 3.84 & 2.01$\times$ & 3.53 & 1.83$\times$ \\
 & SpecVLM & 3.05 & 2.14$\times$ & 3.42 & 2.11$\times$ & 3.02 & 1.89$\times$ & 3.29 & 2.12$\times$ & 3.48 & 2.14$\times$ & 3.49 & 2.10$\times$ \\
 & \blue{\method} & \blue{\textbf{10.29}} & \blue{\textbf{2.71$\times$}} & \blue{\textbf{6.29}} & \blue{\textbf{2.31$\times$}} & \blue{\textbf{8.28}} & \blue{\textbf{2.43$\times$}} & \blue{\textbf{8.44}} & \blue{\textbf{2.40$\times$}} & \blue{\textbf{7.19}} & \blue{\textbf{2.54$\times$}} & \blue{\textbf{9.25}} & \blue{\textbf{2.44$\times$}} \\
\midrule

\multirow{5}{*}{\shortstack[l]{LLaVA-OneVision\\ \qquad 7\&72B}} 
  & Vanilla SD & 4.01 & 1.88$\times$ & 3.32 & 1.89$\times$ & 3.28 & 1.82$\times$ & 3.17 & 1.73$\times$ & 2.87 & 1.64$\times$ & 2.94 & 1.68$\times$ \\
 & SD-tree & 4.47 & 2.03$\times$ & 3.68 & 2.08$\times$ & 3.63 & 2.02$\times$ & 3.57 & 2.08$\times$ & 3.30 & 1.90$\times$ & 3.13 & 1.93$\times$ \\
 & SpecVLM & 4.32 & 2.10$\times$ & 3.48 & 2.68$\times$ & 3.40 & 2.35$\times$ & 3.33 & 2.37$\times$ & 2.96 & 2.00$\times$ & 3.32 & 2.37$\times$ \\
 & \blue{\method} & \blue{\textbf{8.93}} & \blue{\textbf{3.27$\times$}} & \blue{\textbf{8.05}} & \blue{\textbf{3.31$\times$}} & \blue{\textbf{7.93}} & \blue{\textbf{3.51$\times$}} & \blue{\textbf{8.97}}& \blue{\textbf{3.25$\times$}} & \blue{\textbf{7.96}} & \blue{\textbf{3.17$\times$}}& \blue{\textbf{8.39}} & \blue{\textbf{3.16$\times$}} \\
\midrule

\multirow{5}{*}{\shortstack[l]{Qwen3-VL\\\quad 7\&32B}} 
  & Vanilla SD & 3.39 & 1.35$\times$ & 3.79 & 1.27$\times$ & 3.84 & 1.32$\times$ & 3.62& 1.21$\times$ & 3.67 & 1.23$\times$ & 3.88 & 1.39$\times$ \\
 & SD-tree & 3.91 & 1.73$\times$ & 4.25 & 1.63$\times$ & 4.25 & 1.72$\times$ & 4.51 & 1.69$\times$ & 2.51 & 1.53$\times$ & 4.51 & 1.88$\times$ \\
 & SpecVLM & 3.87 & 1.89$\times$ & 4.01 & 2.05$\times$ & 4.12 & 2.03$\times$ & 4.13 & 1.96$\times$ & 2.33 & 1.85$\times$ & 4.20 & 2.11$\times$ \\
 & \blue{\method} & \blue{\textbf{4.67}} & \blue{\textbf{2.38$\times$}} & \blue{\textbf{4.91}} & \blue{\textbf{2.35$\times$}} & \blue{\textbf{4.64}} & \blue{\textbf{2.30$\times$}} & \blue{\textbf{5.91}} & \blue{\textbf{2.24$\times$}} & \blue{\textbf{2.82}} & \blue{\textbf{2.15$\times$}} & \blue{\textbf{5.39}} & \blue{\textbf{2.32$\times$}} \\
\bottomrule
\end{tabular}%
}
\end{table*}

\subsection{Experiment Setups} 

\paragraph{Tasks and Datasets.} We evaluate \method on four video-LLMs: video-SALMONN2+~\citep{videosalmonn2}, Qwen2.5-VL~\citep{qwen25vl}, LLaVA-OneVision~\citep{llavaov}, and Qwen3-VL~\citep{qwen3vl}. We use caption and understanding datasets, including Video-MME~\citep{videomme}, VideoDetailCaption (VDC)~\citep{vdc}, MVBench~\citep{mvbench}, LongVideoBench~\citep{longvideobench}, MLVU~\citep{mlvu}, and LVBench~\citep{lvbench}.

  \paragraph{Implementations and Metrics.}
We conduct experiments on four NVIDIA H200 (140GB) GPUs with batch size of 1. We employ 7B models as draft models and 32B or 72B models as target models. We use two evaluation metrics: (i) wall-time speedup relative to standard auto-regressive decoding, and (ii) mean accepted tokens (MAT). MAT accumulates accepted tokens across verification rounds, including consecutive rounds where all $\gamma$ tokens are successfully verified. Output quality is not evaluated since our acceleration is lossless. More details are provided in Appendix \ref{app:impl}.

\subsection{Main Results}

We conduct extensive experiments on the aforementioned benchmarks. As shown in Table~\ref{tab:merged_performance}, \method consistently outperforms Vanilla SD, SD-tree, and SpecVLM across all backbone architectures and datasets, demonstrating robustness at varying model scales. Notably, \method achieves an inference speedup of up to $3.51\times$ compared to vanilla auto-regressive baselines. Moreover, \method yields significantly higher MAT compared to existing baselines. Specifically, when utilizing video-SALMONN2+ as the backbone model on the MLVU dataset, \method achieves a peak MAT of $12.42$. This not only demonstrates the superior alignment between our pruned draft model and the target model but also validates that our parallel SD algorithm effectively mitigates the non-negligible inference overhead associated with draft models.

\subsection{Ablation Study} \label{sec:ab}

\begin{figure}[t]
    \centering
    \includegraphics[width=\columnwidth]{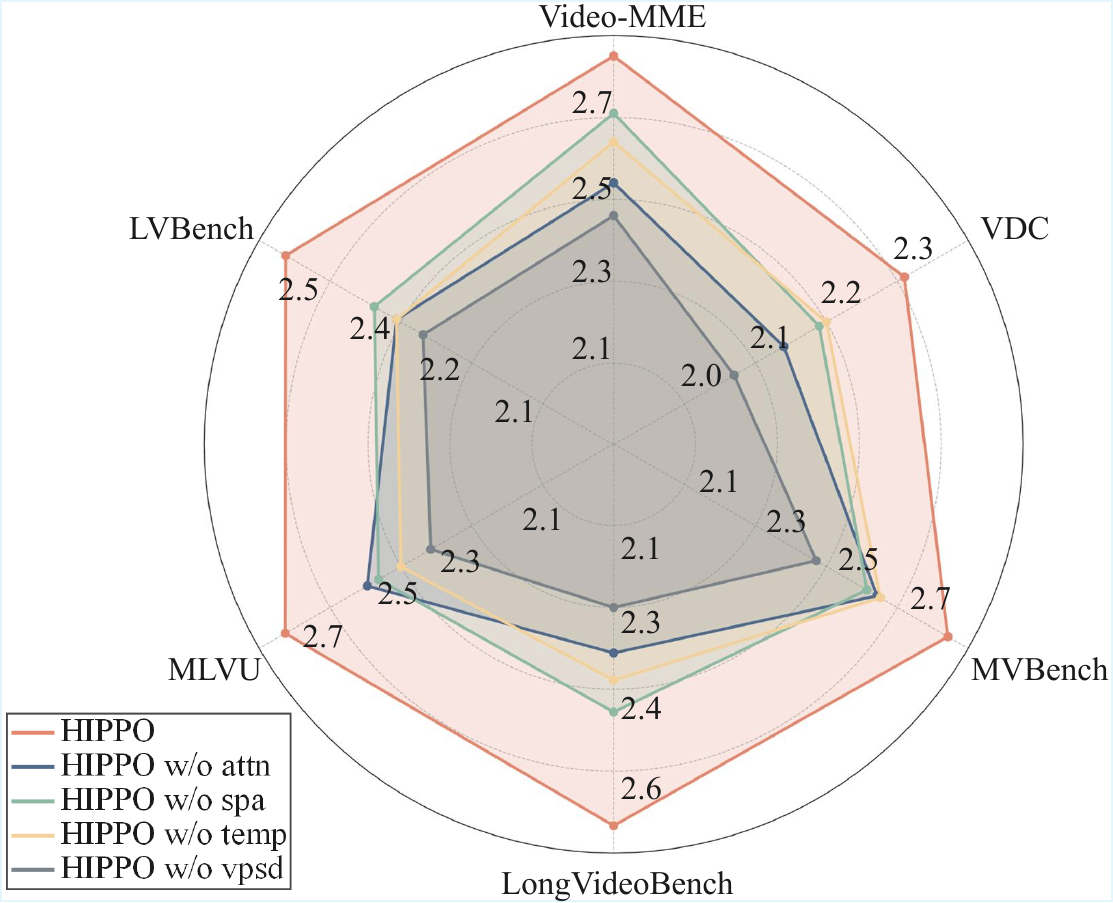}
    \caption{Ablation of \method across six video benchmarks, using video-SALMONN2+ as the backbone.}
    \label{figs:abla}
\end{figure}
To investigate the contribution of each component within \method, we conduct an ablation study in Figure~\ref{figs:abla} using video-SALMONN2+. Results for the other three video-LLMs are in Appendix \ref{app:abla}. We denote \method without global attention score as \method$_{\text{w/o} \hspace{1mm} \text{attn}}$, without temporal redundancy score as \method$_{\text{w/o} \hspace{1mm} \text{temp}}$, without spatial redundancy score as \method$_{\text{w/o} \hspace{1mm} \text{spa}}$, and without video parallel speculative decoding as \method$_{\text{w/o} \hspace{1mm} \text{vpsd}}$.

As shown in Figure~\ref{figs:abla}, the absence of any component within \method results in performance degradation. Among the different scoring components, we observe that utilizing the target model's attention is consistently effective. Additionally, spatiotemporal-aware pruning contributes to improved performance. Specifically, on the MLVU dataset, removing temporal redundancy results in a marked decline in speedup. This underscores that reliance on target model attentions alone cannot effectively capture the essential information for complex video tasks, limiting the speedup ratios.

\subsection{Case Study}

\begin{figure}[htbp]
    \centering
    \includegraphics[width=\columnwidth]{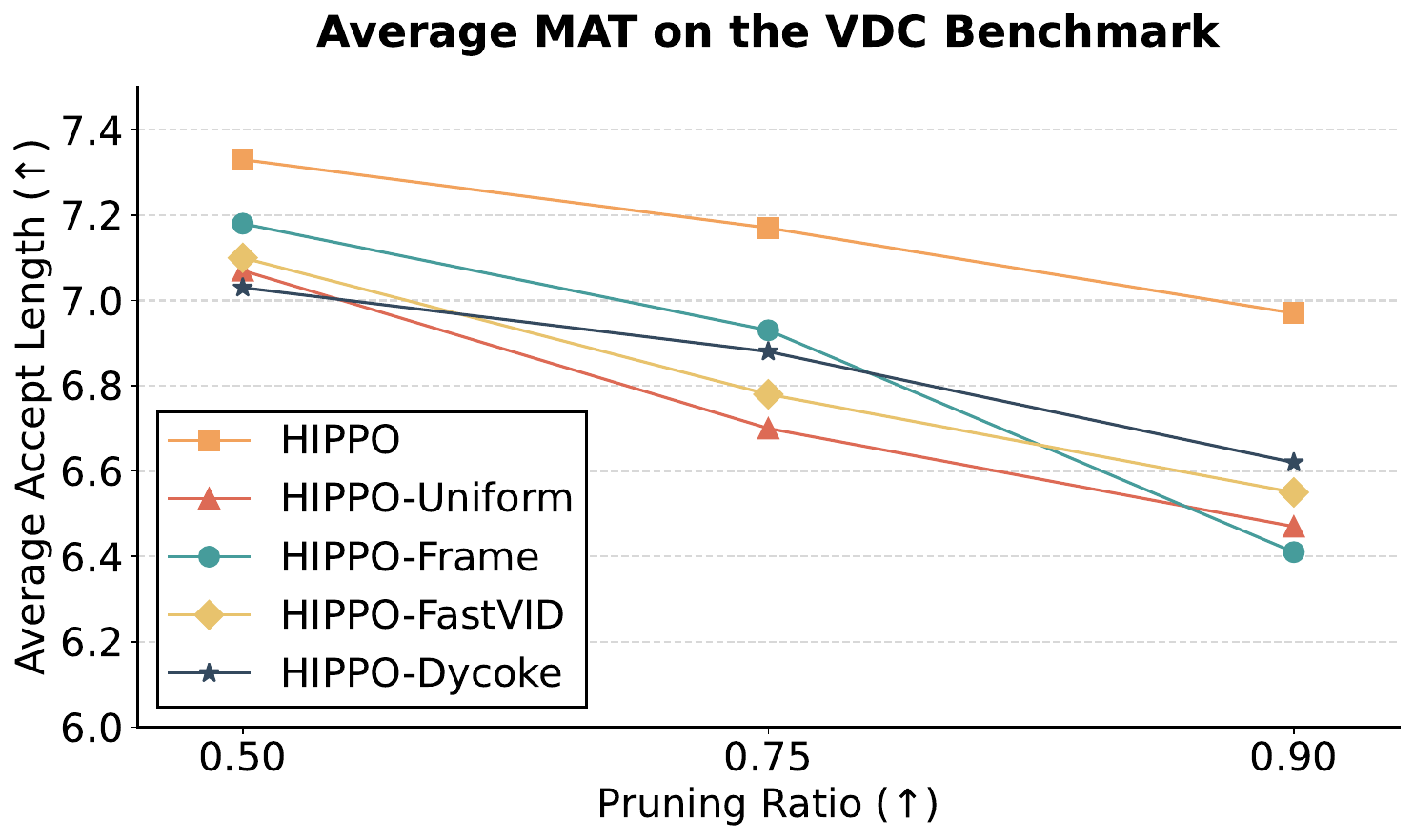}
    \caption{Comparison of different types of pruning methods under various pruning ratios.}
    \label{figs:mat}
\end{figure}
  \paragraph{Exploration of Diverse Pruning Criteria.} \label{sec:prun}
    Numerous studies investigate vision-centric pruning on standalone models to accelerate speed without significantly compromising accuracy~\citep{fastvid,dycoke}. Following this paradigm, we adapt these methods for video-SALMONN2+ 7/72B to establish the following baselines:

    \begin{itemize}
    \item[(1)] \method-Random: randomly sampling top-k video tokens.
    \item[(2)] \method-Frame: full-frame dropping at regular temporal intervals.
    \item[(3)] \method-FastVID: frame-level pruning based on top-k similarity of consecutive frame transitions, following FastVID~\citep{fastvid}.
    \item[(4)] \method-DyCoke: token-level temporal merging adapted from DyCoke~\citep{dycoke}.
    \end{itemize}

    Figure~\ref{figs:mat} demonstrates that vision-centric pruning methods relying solely on a single model consistently underperform compared to \method, which incorporates position-debiased guidance from the target model. We attribute this to two key factors: the draft model benefits from preserving the holistic temporal structure and distribution of the video, and the inclusion of target model signals provides essential guidance for achieving superior alignment. It is worth noting that single-model pruning techniques are not inherently incompatible with \method. Exploring how to leverage intrinsic visual redundancy to refine pruning while integrating information from the target model is a promising direction for future research toward constructing better-aligned draft models.

  \begin{figure}[htbp]
    \centering
    \includegraphics[width=\columnwidth]{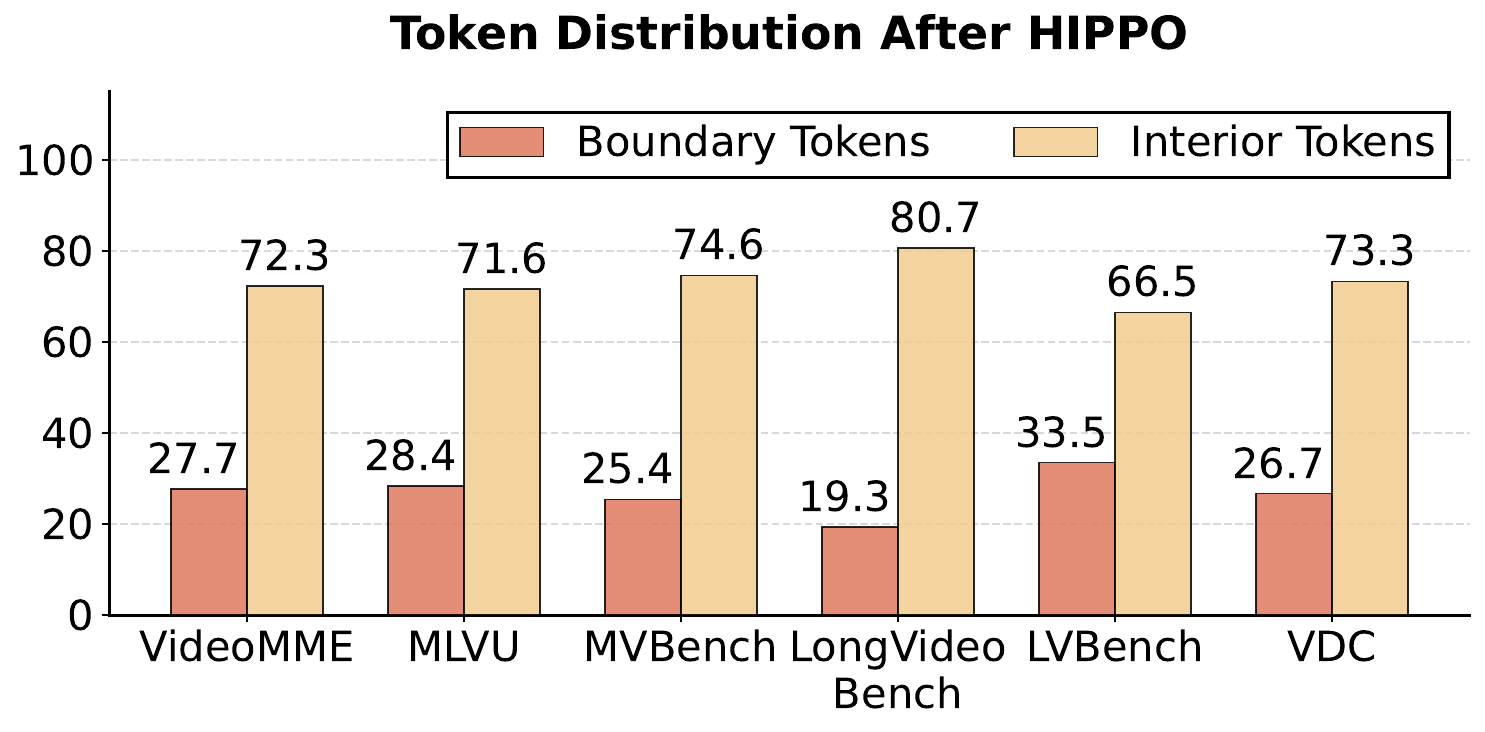}
    \caption{ Per-frame token distributions after \method.}
    \label{figs:position_hippo}
\end{figure}

 \paragraph{Exploration of Token Distributions after \method.}
 As discussed in Section~\ref{sec:motivation}, pruning strategies that rely solely on the target model's attention often suffer from positional bias. To validate the effectiveness of \method, we visualize token distributions after \method, using video-SALMONN2+ as the backbone. As shown in Figure \ref{figs:position_hippo}, tokens selected by \method effectively mitigate this positional bias.
     Within each frame, \method alleviates the tendency to disproportionately retain spatial boundary tokens and focuses on semantic subjects.

\section{Conclusion}
\label{sec:conclusion}

In this paper, we introduced \method, a novel \textbf{h}olist\textbf{i}c-aware \textbf{p}arallel s\textbf{p}eculative dec\textbf{o}ding framework designed to accelerate video-LLM inference.
  \method addresses the efficiency bottlenecks of existing methods through two key innovations:
  (i) a semantic-aware token preservation strategy that integrates \textbf{global attention scores} with \textbf{local visual semantics}, thereby maintaining high alignment between the draft and target models at high pruning ratios;
  (ii) a video parallel speculative decoding algorithm that decouples draft generation from verification, effectively eliminating the mutual waiting overhead.
  Experiments on four video-LLMs across six datasets demonstrate the effectiveness of our approach, achieving up to $3.51\times$ speedup compared to vanilla auto-regressive decoding.

  \section{Limitations}

  We consider a few limitations and future directions.
(i) Specialized draft model training. Our current evaluation employs existing smaller models from the same family as draft models to accelerate the target model. Training lightweight, target-specific draft models with better alignment represents a promising direction for further improving speculative decoding efficiency. Our method is naturally compatible with such specialized draft models, requiring only one-time pruning of the draft model's KV cache during the prefill stage.  As training lightweight video-LLMs remains an emerging area, future work can explore draft model optimization techniques—such as knowledge distillation or architecture specialization—to further enhance speculative decoding efficiency. (ii) Applicability scope. Our method is primarily effective in memory bandwidth scenarios as our adaptive pruning strategy directly targets memory I/O overhead reduction,  extending the approach to high-throughput processing requires further investigation. (iii) Scalability to large batch sizes. Our evaluation focuses on latency-critical scenarios with batch size 1, the primary use case for real-time interaction. For high-throughput scenarios with large batches, GPU compute saturation may occur, and maintaining diverse dynamic tree structures per request could introduce non-trivial overhead.

  \section{Ethical Considerations}

  The data we collect in specialized domains is publicly available and viewable online. The data owners have indicated that the data can be used for scientific research or have not indicated that the data cannot be used for scientific research, and our collection process is also in compliance with regulations. Moreover, there is no unique identification of individuals or offensive content in these data. 

\clearpage

\clearpage
\appendix
\section{More Discussion of Related Work} \label{app:related}

In Section \ref{sec:related}, we briefly discuss the recent achievements of speculative decoding and visual token reduction methods. In this section, we provide a more detailed discussion on existing multi-modal LLMs and their computational challenges, various speculative decoding methods for LLMs, and visual token reduction methods.

\subsection{MLLMs and Computational Challenges}

Recent MLLMs typically adhere to a paradigm that bridges multi-modal encoders with LLMs via specialized modality adapters~\citep{gemini, llava, improved}. Built upon open-source foundations such as the LLaMA series~\citep{llama1,llama2,llama3}, state-of-the-art MLLMs~\citep{gemini, gpt4} have demonstrated remarkable adaptability across diverse visual understanding tasks, significantly enhancing their ability to interpret complex real-world environments.
In the video domain, video-LLaVA~\citep{videollava} aligns image and video adapters to learn unified representations. ShareGPT4Video~\citep{sharegpt4video} utilizes GPT-4 to generate dense video captions for data quality improvement, while LLaVA-Hound~\citep{llavahound} introduces Direct Preference Optimization (DPO) to refine video comprehension capabilities. Furthermore, LLaVA-OneVision~\citep{llavaov} achieves robust performance in single-image, multi-image, and video scenarios, facilitating effective transfer learning from static images to dynamic video content.
Despite these advancements, the integration of extensive visual tokens introduces substantial computational overhead. While visual perception is improved, existing MLLMs continue to suffer from certain perceptual limitations~\citep{eyes, marvel}. Strategies to mitigate these deficiencies by increasing the resolution of input images or videos~\citep{feast, llavauhd} inadvertently exacerbate computational costs. For instance, LLaVA-OneVision~\citep{llavaov} encodes a single video frame into $196$ tokens; consequently, processing a two-minute video at $60$ FPS yields over one million tokens. Such massive visual inputs become a significant portion of the LLM's context window. In this work, we conduct experiments on these representative models to empirically validate the applicability and efficiency of our \method.

\subsection{Speculative Decoding for LLMs}

Pioneering approaches in speculative decoding~\citep{sd1, sd2} typically employ an independent, lightweight draft model to propose short token sequences, which are subsequently verified in parallel by the target model. To ensure reliable speculation, these early methods often utilize existing smaller LLMs as drafters. Alternatively, self-speculative frameworks~\citep{swift, knn} leverage intermediate layers of the target model itself to generate predictions, thereby eliminating the need for auxiliary models. Contemporary research has largely focused on enhancing draft model efficiency; for instance, recent work~\citep{glide} introduces specially trained drafters with reduced latency to maximize the speedup ratio. Several studies have also adapted speculative decoding methods from the text domain to the single-image domain~\citep{msd,msd2, HiViS}.

  Beyond standard decoding, several SD methods address the specific challenges of long-context text inference~\citep{longspec, triforce, magicdec}. These approaches aim to mitigate the Key-Value (KV) cache bottleneck inherent in conventional SD by integrating KV cache sparsification techniques~\citep{adakv}, often utilizing StreamingLLM-style cache management~\citep{attensink} to constrain the draft model's memory budget. However, the absence of visual modality awareness precludes these methods from operating effectively on video tokens. Unlike discrete linguistic tokens, video tokens exhibit substantial redundancy and high temporal correlation. Consequently, the naive application of StreamingLLM or sliding window mechanisms restricts the draft model's attention to a narrow fraction of uncompressed visual content, causing it to fail in capturing the visual semantics of the video.

\subsection{Visual Token Reductions}

In MLLMs, visual redundancy identification facilitates the distillation of visual tokens with high informativeness for faster inference. There are two main research directions: (i) Vision-centric strategies analyze the image's structure and feature distribution to discard less relevant visual tokens~\citep{llvavid, cls, AdaTP}. Approaches such as LLaMA-VID~\citep{llvavid} and DeCo~\citep{deco} address this by modifying model architectures with additional training; however, this inevitably escalates computational expenses. While ToMe~\citep{tome} achieves token reduction without retraining, it risks disrupting early-stage cross-modal interactions~\citep{pdrop}. LLaVA-PruMerge~\citep{prumerge} selectively retains salient tokens while merging less critical ones based on key similarity. (ii) Instruction-centric strategies typically use cross-modal attention analysis or gradient accumulation to identify redundant tokens~\citep{multi, less}. Tokens with low attention or negligible gradient impact are deemed redundant~\citep{zipvl}. Building on this, some studies explore learned importance scoring and retain the highlighted visual tokens~\citep{fopru}. Distinct from these single-model pruning approaches, \method focuses on leveraging signals derived from the target model to guide draft model pruning. Consequently, our \method is orthogonal to existing techniques; the pruned draft model can be seamlessly combined with these methods to adapt to specific downstream tasks.

\begin{figure*}[t]
      \centering
      \begin{minipage}[b]{0.48\textwidth}
          \centering
          \includegraphics[width=\columnwidth]{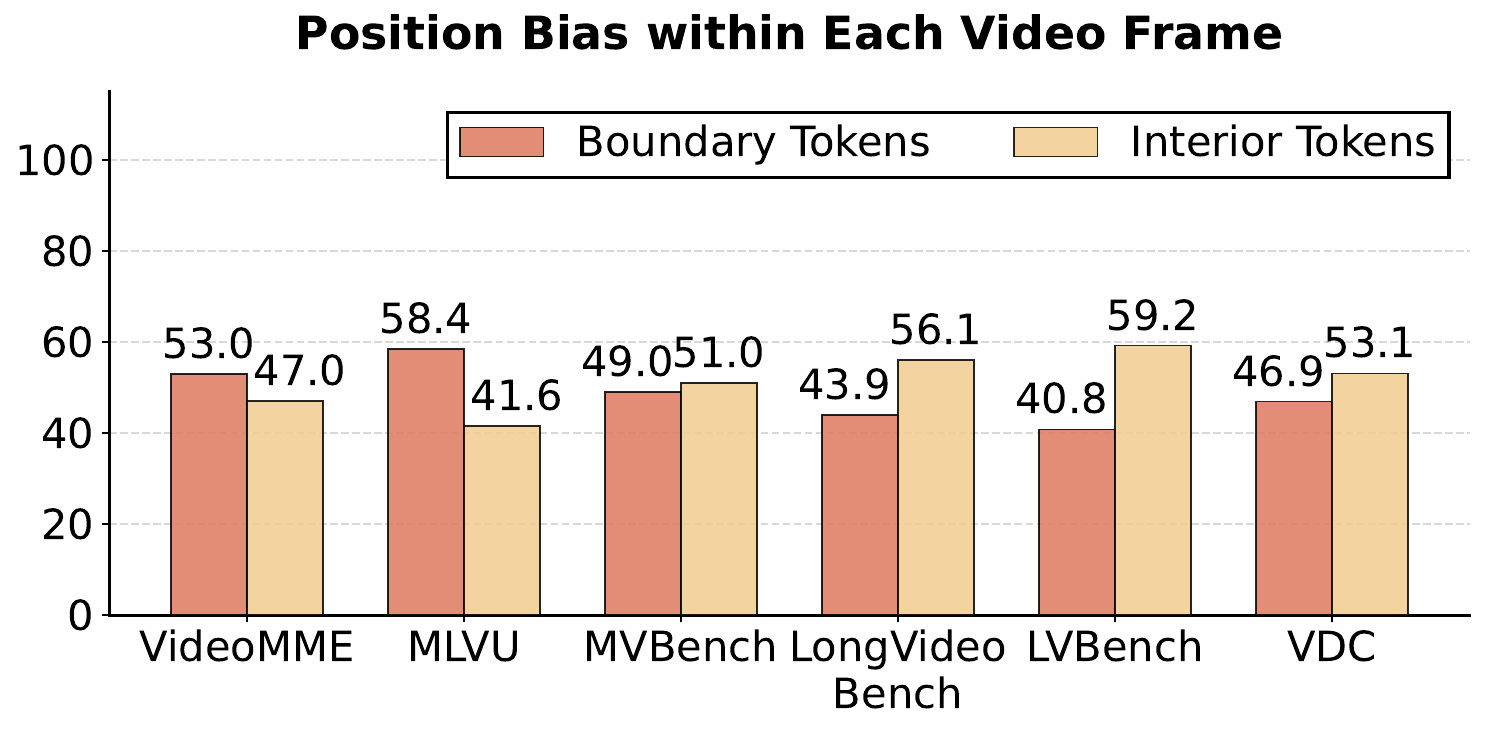}
          \caption{Statistics of per-frame positional bias using Qwen2.5-VL as the backbone model.}
          \label{figs:position_bias_qwen25}
      \end{minipage}
      \hfill
      \begin{minipage}[b]{0.48\textwidth}
          \centering
          \includegraphics[width=\columnwidth]{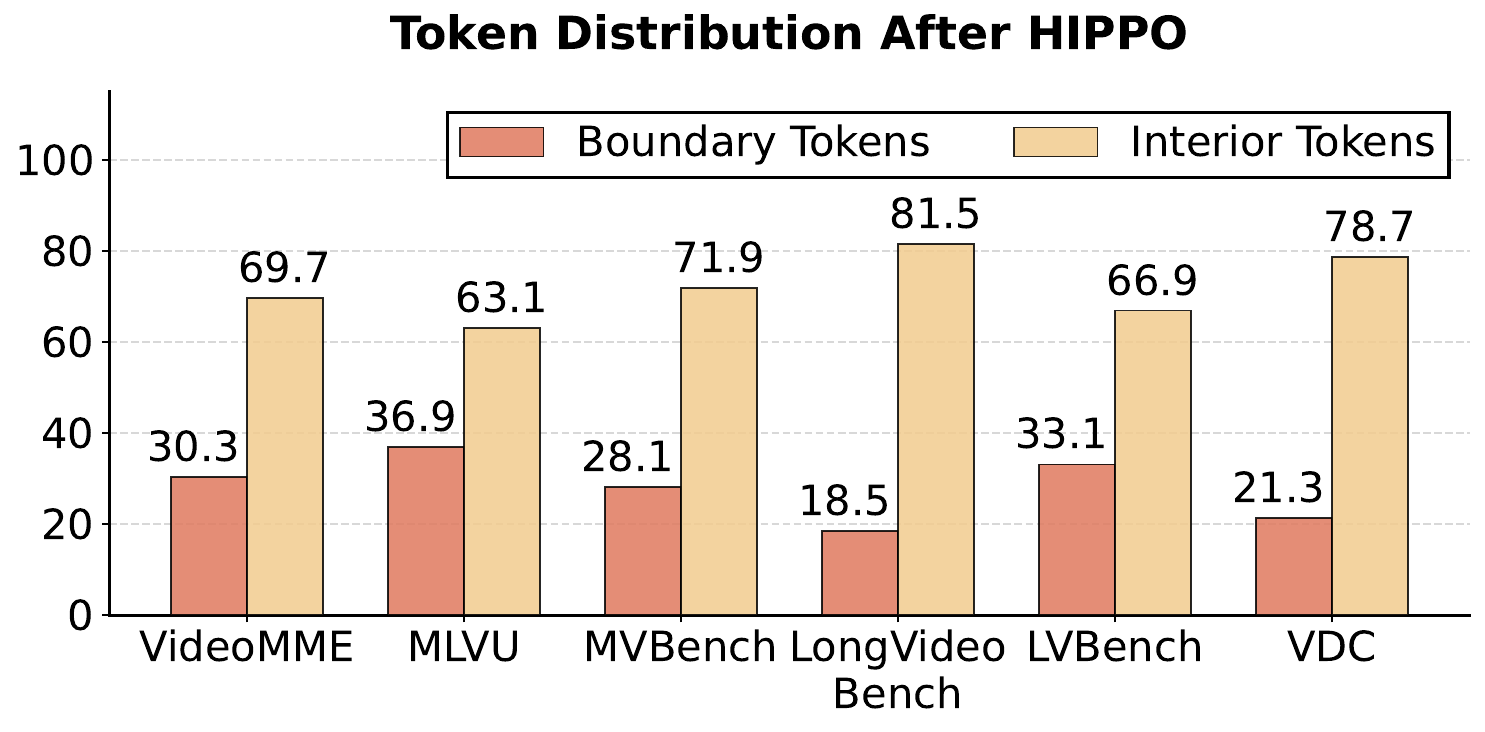}
          \caption{Per-frame token distributions after \method using Qwen2.5-VL as the backbone model.}
          \label{figs:position_hippo_qwen25}
      \end{minipage}
  \end{figure*}

\begin{figure*}[t]
      \centering
      \begin{minipage}[b]{0.48\textwidth}
          \centering
          \includegraphics[width=\columnwidth]{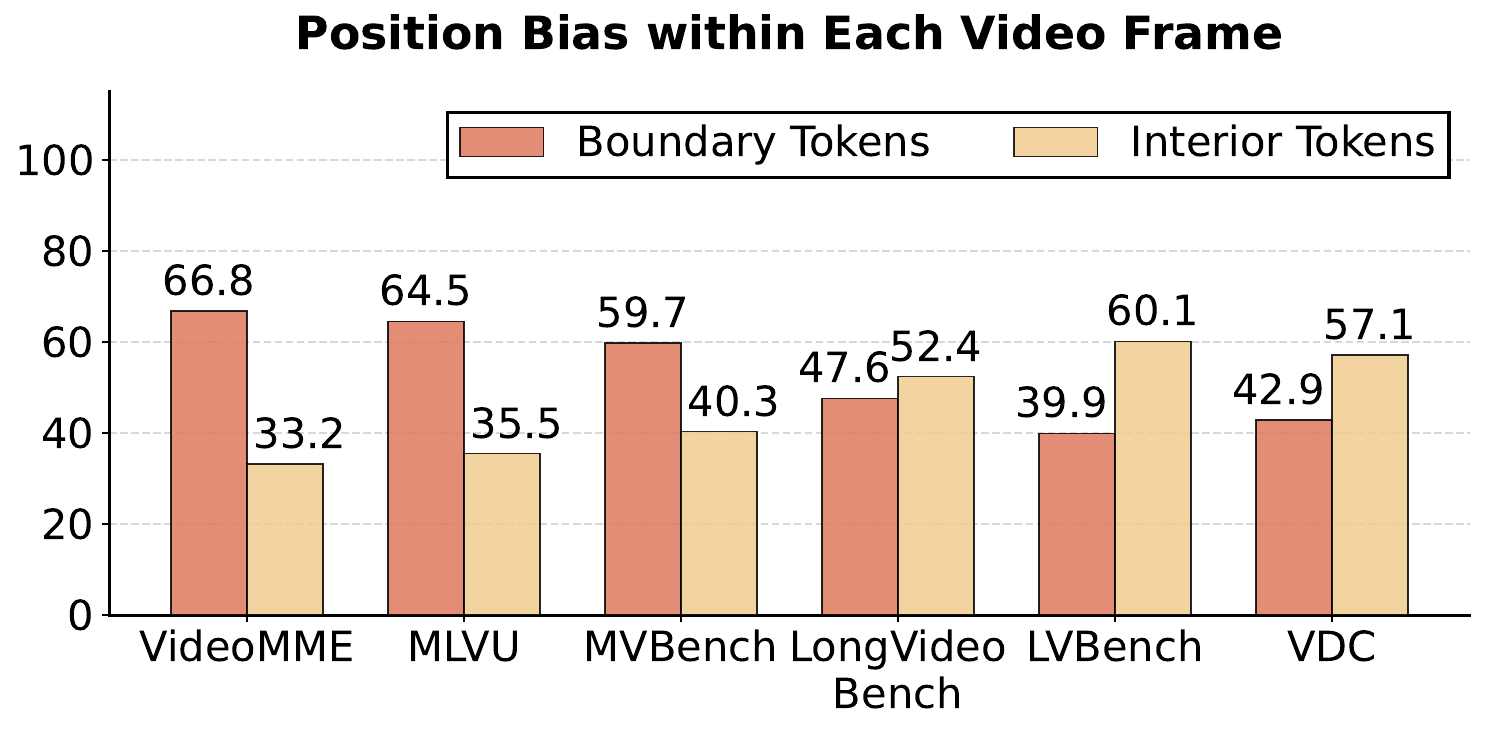}
          \caption{Statistics of per-frame positional bias using LLaVA-OneVision as the backbone model.}
          \label{figs:position_bias_llavaov}
      \end{minipage}
      \hfill
      \begin{minipage}[b]{0.48\textwidth}
          \centering
          \includegraphics[width=\columnwidth]{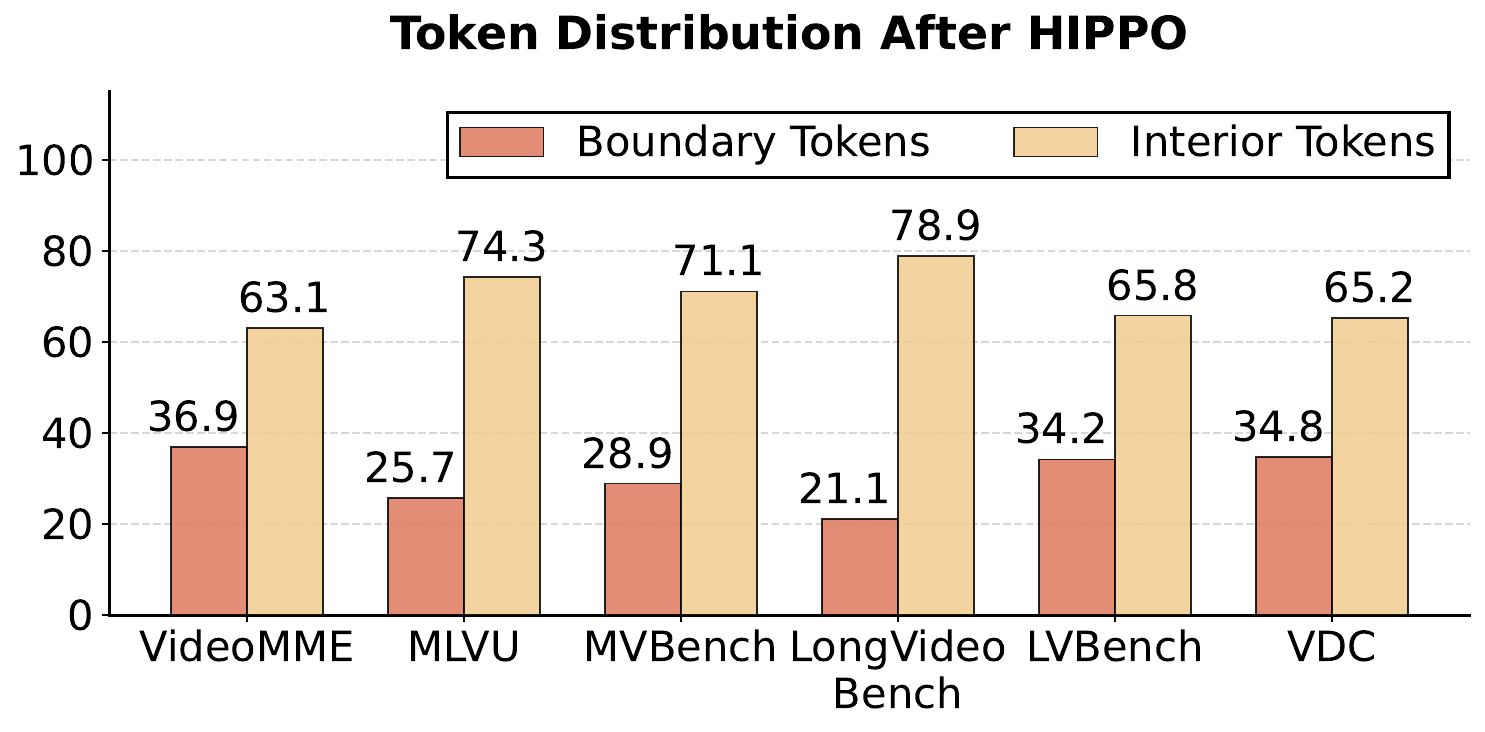}
          \caption{Per-frame token distributions after \method using LLaVA-OneVision as the backbone model.}
          \label{figs:position_hippo_llavaov}
      \end{minipage}
  \end{figure*}

\section{More Details of Experiment Setups}\label{app:impl}

We present more details of datasets and experiment setups in this section.


  \paragraph{Details of Baselines.}

    We implement and evaluate three inference acceleration baselines to benchmark our approach:
    \textbf{(i) Vanilla Speculative Decoding (SD)}: we employ a compact video-LLM as the draft model to generate preliminary token sequences, which are subsequently verified by the target model. This foundational approach establishes the basic speculative decoding framework for video-language tasks, where the draft model produces candidate continuations in an auto-regressive manner, and the target model validates these proposals in parallel.
    \textbf{(ii) Speculative Decoding with Tree Attention (SD-tree)}: Building upon vanilla SD, this method constructs a draft tree structure by proposing multiple candidate tokens at each position, thereby exploring diverse continuation paths simultaneously. The tree-based representation allows the target model to verify multiple speculative branches in parallel, potentially increasing the acceptance rate and overall throughput compared to sequential speculation.
    \textbf{(iii) SpecVLM}: this approach introduces an attention-guided token selection mechanism, where candidate tokens are strategically chosen based on the target model's attention patterns from previous decoding steps. By leveraging these attention weights as guidance signals, SpecVLM prunes redundant or low-probability candidates early in the speculation process, focusing computational resources on more promising token sequences and reducing unnecessary verification overhead.

    Notably, we avoid introducing additional training procedures for the draft models in our baseline implementations. This design decision is motivated by the observation that the primary computational bottleneck in video-LLM inference stems from the management and retrieval of large key-value caches rather than the forward pass through model parameters. Therefore, our baselines focus on optimizing the decoding strategy and speculation efficiency while maintaining the original model weights.

  \paragraph{Details of Experimental Setups.}  We conduct experiments on four NVIDIA H200 (140GB) GPUs. We employ 7B models as draft models and 32B or 72B models as target models. We perform all evaluations with a batch size of 1 to follow the standard settings and simulate real-world latency-critical inference scenarios. Given each prompt, the target model is employed to generate 256 tokens following greedy decoding. When video tokens are pruned, we remove the corresponding video features based on the pruning ratio r. During evaluation, the default pruning ratio r is set to 90\%. The value of $\gamma$ is set to 5, 5, 7, and 3 for video-SALMONN2+ 7B/72B, Qwen2.5-VL 7B/72B, LLaVA-OneVision 7B/72B, and Qwen3-VL 7B/32B, respectively. The crop size $M$ is set to 5 × 5 for all models and benchmarks. To ensure a fair comparison, all methods are evaluated under an identical hardware environment using model parallelism. For the  AR baseline, the target model spans four GPUs. For the all SD-based methods we apply the standard protocol by deploying both the draft and target models across the four GPUs.

  For LLaVA-OneVision, we uniformly sample 64 and 128 frames to generate a 196 × 64 and 196 × 128 video token input, respectively. For Qwen series and video-SALMONN2+, we adjust the FPS to generate input of comparable length. For video-SALMONN2+, we only apply pruning to the visual components. The audio tokens are excluded from the scoring process and are retained in their entirety as input. Our entire
  experimental framework is implemented using \texttt{PyTorch} \citep{pytorch} version \texttt{2.8.0}, a widely-adopted deep learning framework that provides comprehensive support for GPU-accelerated tensor operations and automatic differentiation. We leverage the \texttt{Transformers} \citep{transformers} library version \texttt{4.57.0} from Hugging Face.
    The codebase is compiled and executed with CUDA toolkit version \texttt{12.8}. More details for the best performance of each task and benchmark can be seen within our code.

  \begin{figure*}[t]
      \centering
      \begin{minipage}[b]{0.48\textwidth}
          \centering
          \includegraphics[width=\columnwidth]{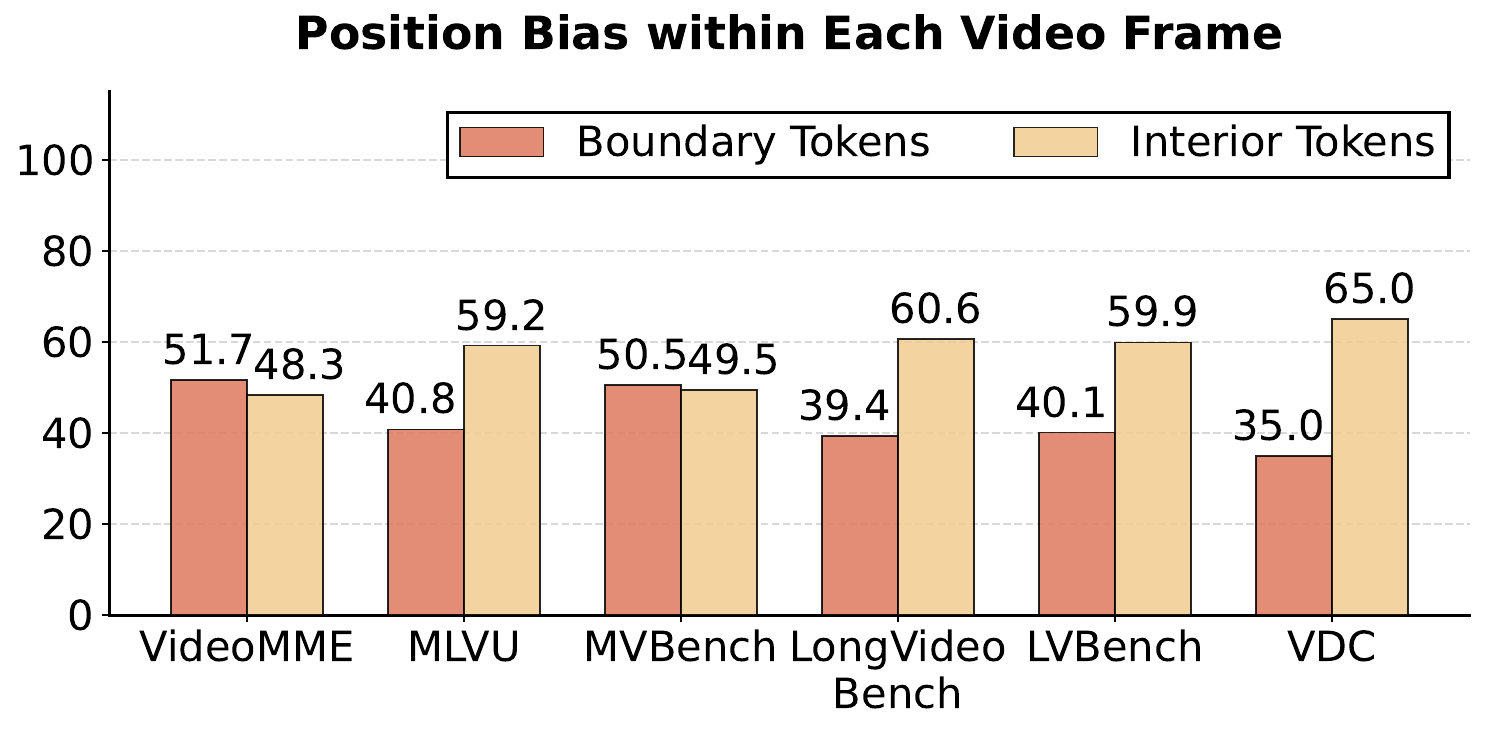}
          \caption{Statistics of per-frame positional bias using Qwen3-VL as the backbone model.}
          \label{figs:position_bias_qwen3vl}
      \end{minipage}
      \hfill
      \begin{minipage}[b]{0.48\textwidth}
          \centering
          \includegraphics[width=\columnwidth]{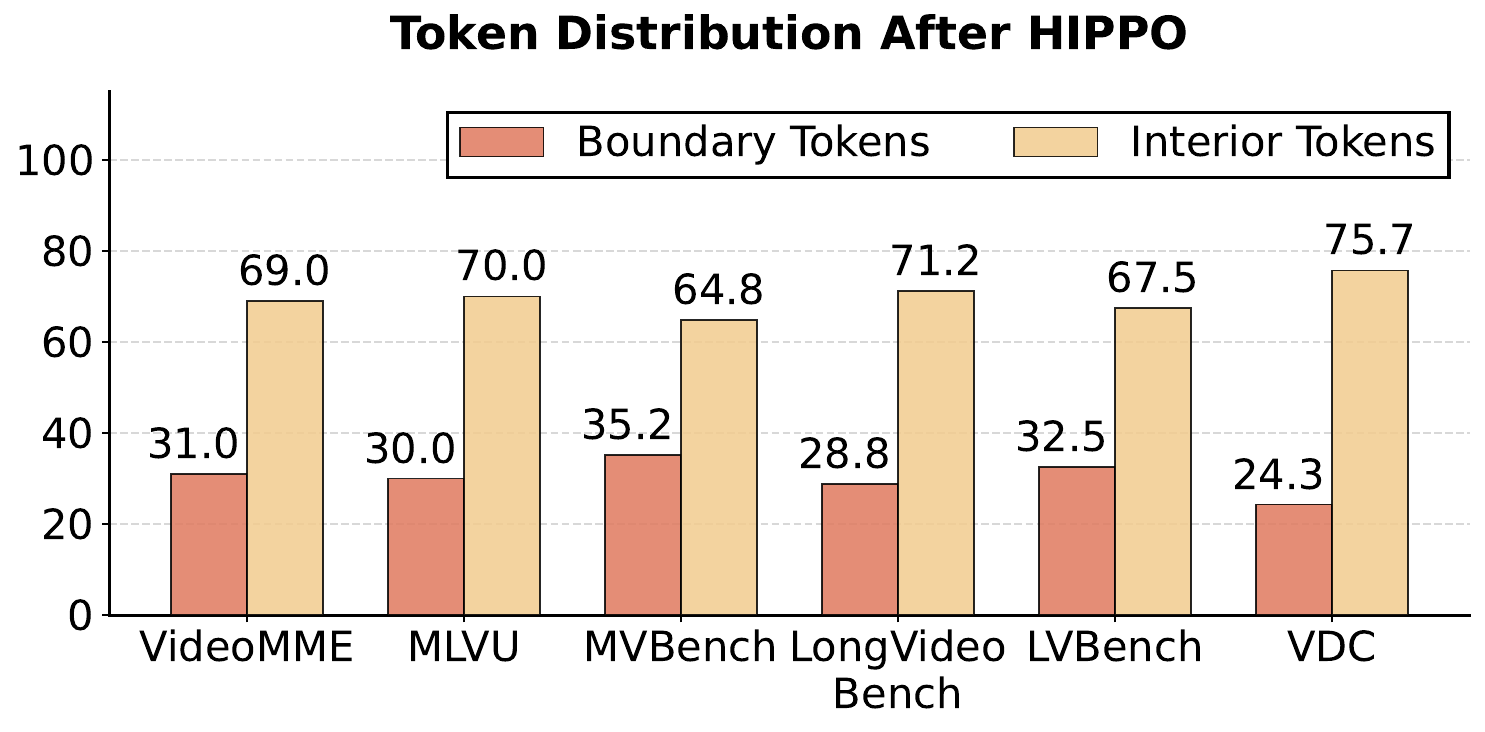}
          \caption{Per-frame token distributions after \method using Qwen3-VL as the backbone model.}
          \label{figs:position_hippo_qwen3vl}
      \end{minipage}
  \end{figure*}

  \begin{table}[t]
  \centering
  \caption{Inference time breakdown (s) of video-SALMONN2+ 7B/72B pair. ``-'' indicates that vanilla AR does not require a draft model or pruning. ``overlapped'' indicates that due to our parallelized HIPPO, the execution time of the draft model is  overlapped by the concurrent execution of the target model.}
  \label{table_breakdown}
  \resizebox{\columnwidth}{!}{%
  \begin{tabular}{lcc}
  \toprule
  \textbf{Operation} & \textbf{Vanilla} & \textbf{HIPPO} \\
  \midrule
  Target Model Prefilling & 33.71 & 33.71 \\
  Target Model Decoding & 131.13 & 25.11 \\
  Draft Model Prefilling & - & 2.79 (overlapped) \\
  Draft Model Decoding & - & 23.77 (overlapped) \\
  Video Token Pruning & - & 0.11 (overlapped) \\
  \midrule
  \textbf{Total Latency} & 164.84 & 58.82 \\
  \bottomrule
  \end{tabular}}
  \end{table}

  \section{Inference Time Breakdown}

  Beyond accelerating the decoding stage of target video-LLMs, \method introduces minimal overhead during the pruning process. We randomly sample $10$ videos in Video-MME, employ a video-SALMONN2+ 7B/72B model pair for evaluation, and report  average wall-time of each operation. As shown in \ref{table_breakdown}, through video token pruning (only 0.11s for the whole decoding process), the prefill length of the draft model is substantially reduced. Furthermore, since \method parallelizes the prefill and decoding processes of both the draft and target model, the execution times of these operations overlap, thereby further reducing the overall latency. 

\section{More Comparison with Parallel Speculative Methods}

\begin{table*}[htbp]
\centering
\caption{Performance comparison with parallel speculative decoding methods, using video-SALMONN2+ 7B/72B as the backbone model pair. We report wall-time speedup relative to standard auto-regressive decoding. We \textbf{bold} the best results for each benchmark.}
\label{tab:app_para_compa}
\resizebox{0.95\textwidth}{!}{%
\setlength{\tabcolsep}{8pt} 
\begin{tabular}{lcccccc}
\toprule
\multirow{2}{*}{\textbf{Method}} & \textbf{VideoMME} & \textbf{VDC} & \textbf{MVBench} & \textbf{LongVideoBench} & \textbf{MLVU} & \textbf{LVbench} \\
 & Spd. & Spd. & Spd. & Spd. & Spd. & Spd. \\
\midrule
PEARL-video & 2.24$\times$ & 1.87$\times$ & 2.45$\times$ & 2.18$\times$ & 2.20$\times$ & 2.21$\times$\\
PEARL w/ Syn-Prefill & 2.46$\times$ & 2.03$\times$ & 2.64$\times$ & 2.35$\times$ & 2.30$\times$ & 2.33$\times$\\
PEARL-SpecVLM & 2.52$\times$ & 2.06$\times$ & 2.69$\times$ & 2.47$\times$ & 2.46$\times$ & 2.44$\times$\\
\blue{\method} & \blue{\textbf{2.85$\times$}} & \blue{\textbf{2.31$\times$}} & \blue{\textbf{2.89$\times$}} & \blue{\textbf{2.74$\times$}} & \blue{\textbf{2.78$\times$}} & \blue{\textbf{2.64$\times$}} \\
\bottomrule
\end{tabular}%
}
\end{table*}

As discussed in the main text, within the textual domain, PEARL~\citep{pearl} employs a parallelization strategy for draft and target models to enable adaptive draft lengths and eliminate mutual waiting latency. Building upon this, we further optimize the framework for the video domain. Specifically, observing that the target model's prefill latency in video-LLMs is dominated by the computationally expensive vision encoder—thereby significantly exceeding the draft model's prefill time—we propose the Synchronous Draft-Target Prefill mechanism. To comprehensively evaluate the effectiveness of our \method, we compare it against several baselines: (i) PEARL-Video, a direct adaptation of the textual PEARL to the video domain; (ii) PEARL w/ Syn-Prefill, which incorporates our proposed synchronous draft-target prefill into the PEARL framework; and (iii) PEARL-SpecVLM, which integrates SpecVLM's attention-based draft selection mechanism within the parallel speculative decoding framework. Using video-SALMONN2+ 7B/72B as the backbone model pair, we present the results in Table \ref{tab:app_para_compa}. We observe that our \method consistently outperforms competing parallel SD methods. This indicates that \method effectively preserves visual semantic tokens, thereby improving draft quality. Furthermore, PEARL w/ Syn-Prefill achieves consistent improvements over vanilla PEARL-video across all benchmarks. This demonstrates that employing synchronous draft-target prefill in the video domain allows for the accumulation of a larger initial buffer of candidate tokens, effectively utilizing the target model's prefilling stage.

\section{More Motivated Experiments}

\begin{figure}[t]
    \centering
    \includegraphics[width=\columnwidth]{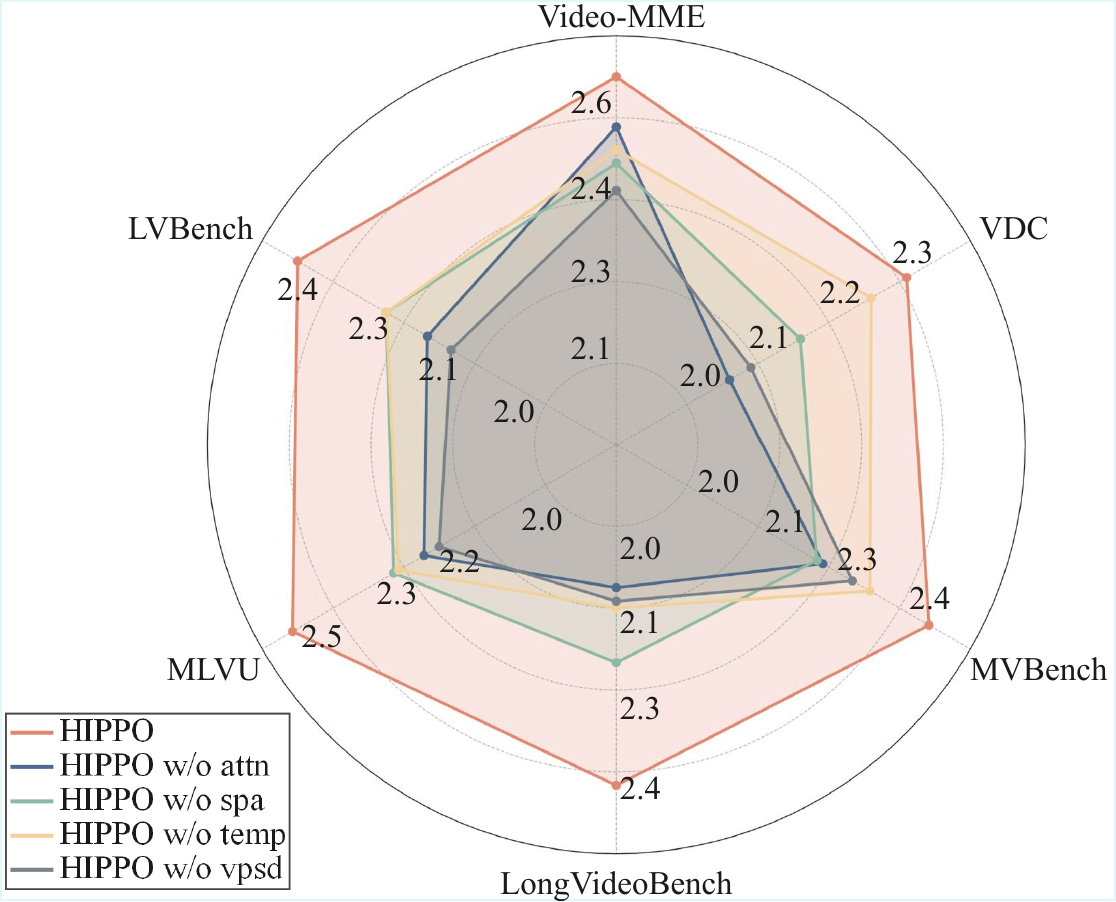}
    \caption{Ablation of \method across six video benchmarks, using Qwen2.5-VL as the backbone.}
    \label{figs:abla_qwen25vl}
\end{figure}

\begin{figure}[t]
    \centering
    \includegraphics[width=\columnwidth]{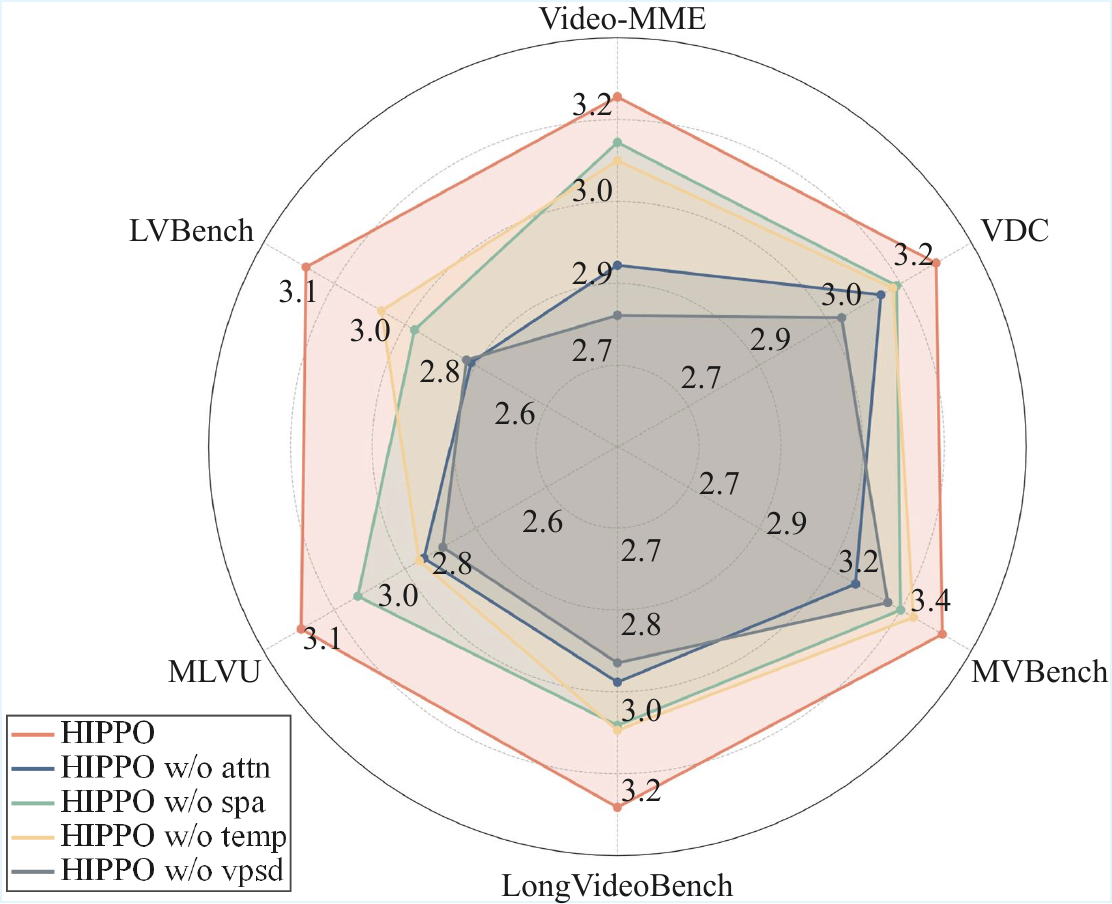}
    \caption{Ablation of \method across six video benchmarks, using LLaVA-OneVision as the backbone.}
    \label{figs:abla_llavaov}
\end{figure}

  \begin{figure*}[t]
      \centering
      \begin{minipage}[b]{0.48\textwidth}
    \centering
    \includegraphics[width=\columnwidth]{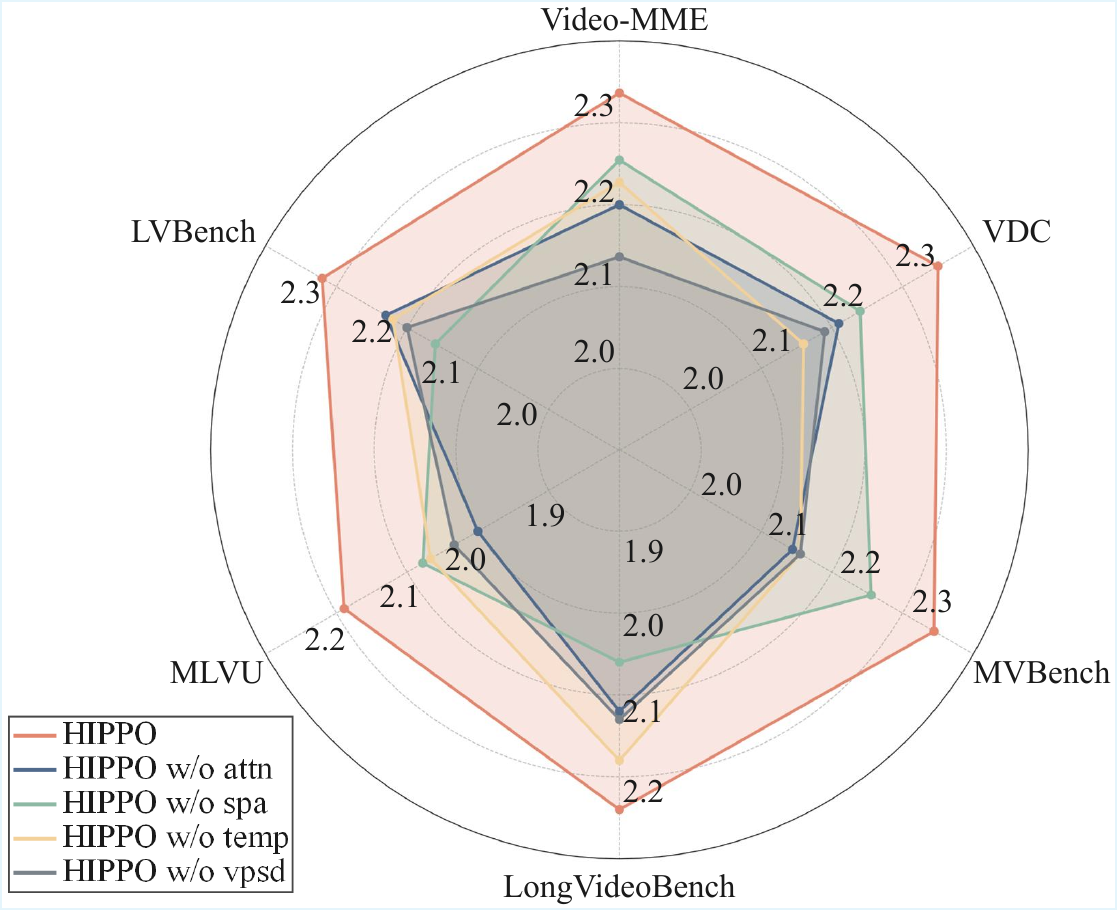}
    \caption{Ablation of \method across six video benchmarks, using Qwen3-VL as the backbone.}
    \label{figs:abla_qwen3vl}
      \end{minipage}
      \hfill
      \begin{minipage}[b]{0.48\textwidth}
    \centering
    \includegraphics[width=\columnwidth]{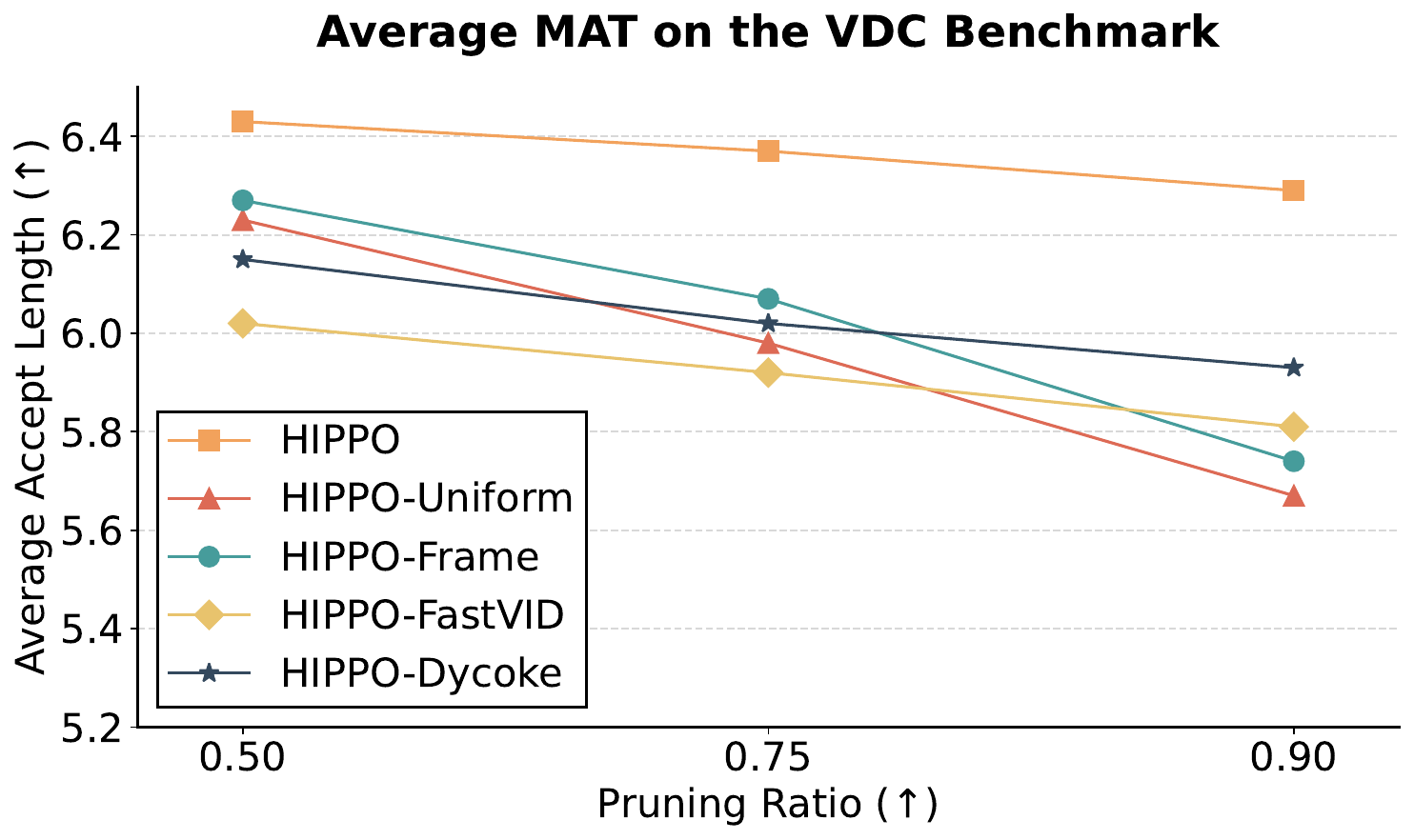}
    \caption{Comparison of different types of pruning methods under various pruning ratios using Qwen2.5-VL as the backbone model.}
    \label{figs:mat_qwen25}
      \end{minipage}
  \end{figure*}

  \begin{figure*}[t]
      \centering
      \begin{minipage}[b]{0.48\textwidth}
       \centering
    \includegraphics[width=\columnwidth]{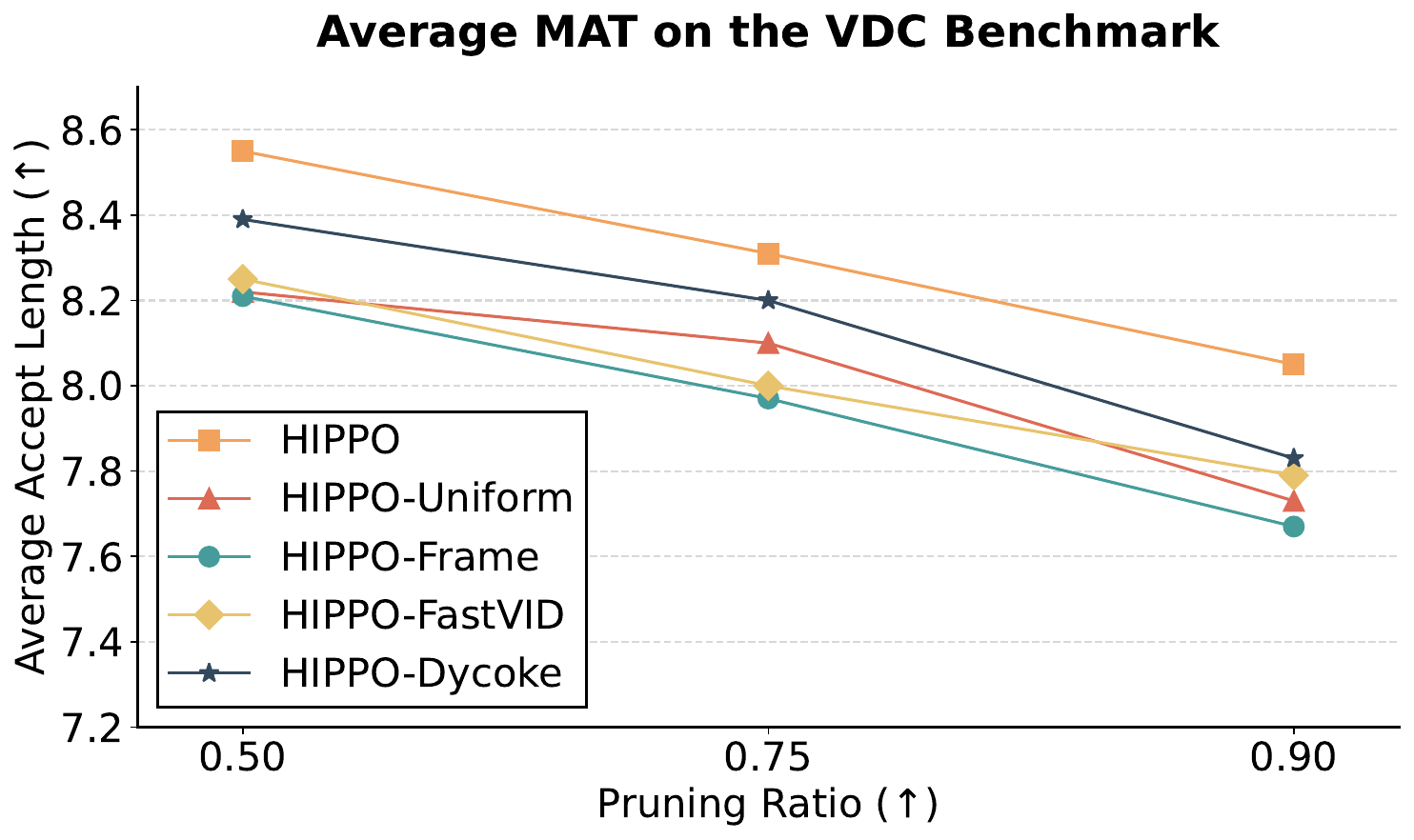}
    \caption{Comparison of different types of pruning methods under various pruning ratios using LLaVA-OneVision as the backbone model.}
    \label{figs:mat_llavaov}
      \end{minipage}
      \hfill
      \begin{minipage}[b]{0.48\textwidth}
       \centering
    \includegraphics[width=\columnwidth]{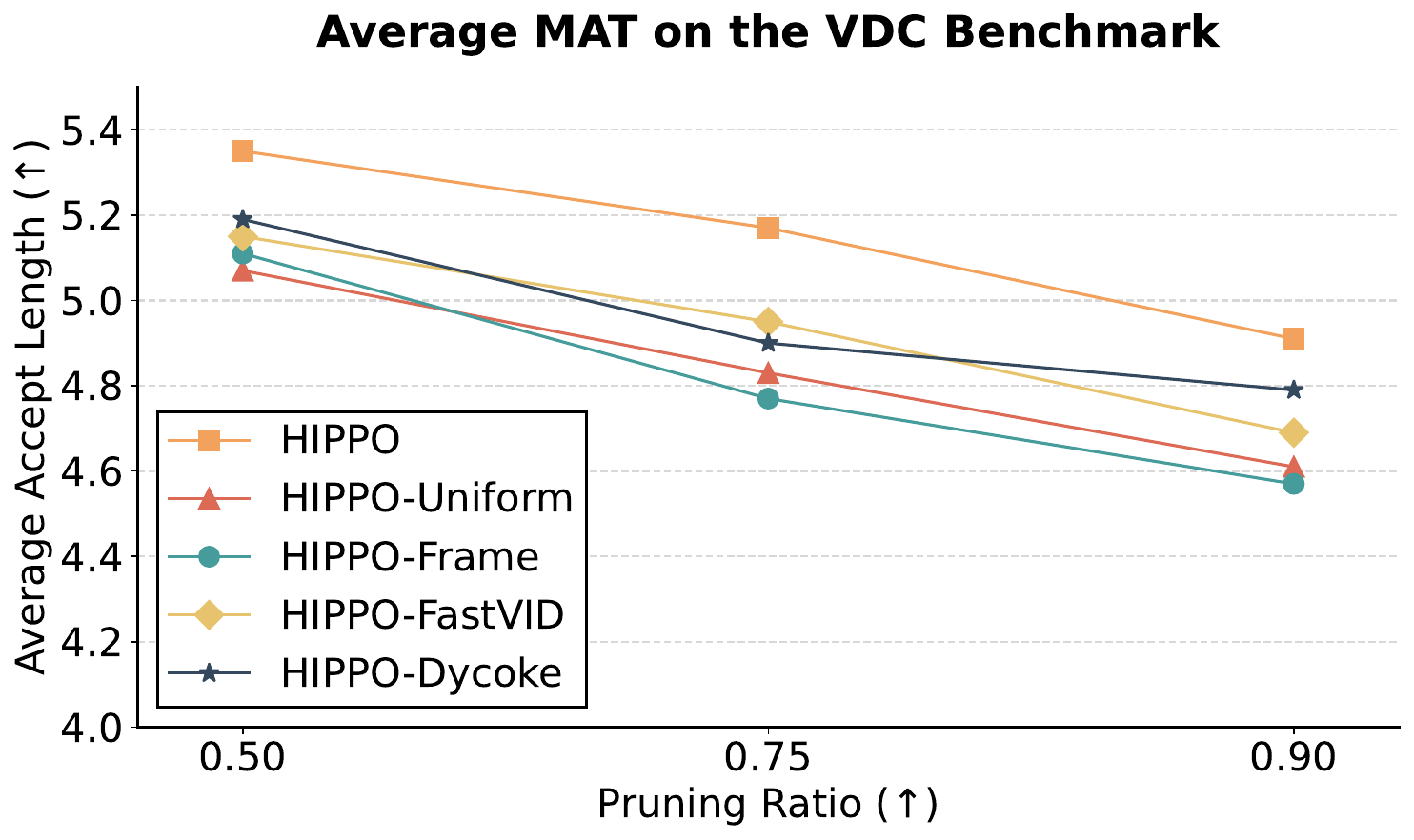}
    \caption{Comparison of different types of pruning methods under various pruning ratios using Qwen3-VL as the backbone model.}
    \label{figs:mat_qwen3}
      \end{minipage}
  \end{figure*}

   In Section \ref{sec:motivation}, we present comprehensive position bias statistics utilizing video-SALMONN2+ as the backbone model. To establish the generality and broader applicability of our observations, we conduct extensive empirical analyses across a diverse range of video-LLMs with varying architectures and capacities. These systematic experiments substantiate that the identified inefficiencies associated with position bias are not merely model-specific phenomena, but rather represent fundamental and inherent challenges endemic to attention-based pruning mechanisms in existing video-LLM SD approaches.

  As illustrated in Figures \ref{figs:position_bias_qwen25}, \ref{figs:position_bias_llavaov}, and \ref{figs:position_bias_qwen3vl}, when employing solely the target model's attention scores to extract highlighted visual tokens across different video-LLM architectures, a pronounced position bias consistently emerges across all evaluated models. 
Conversely, as shown in Figures \ref{figs:position_hippo_qwen25}, \ref{figs:position_hippo_llavaov}, and \ref{figs:position_hippo_qwen3vl}, we observe that tokens selected by our proposed \method effectively mitigate this positional bias phenomenon. \method preserves semantically salient tokens distributed throughout frames, rather than exhibiting the tendency to disproportionately favor tokens located at spatial boundaries, thereby achieving a balanced token selection strategy that captures the true informational content of the video.

\section{More Results of Ablation Study} \label{app:abla}

  In Section \ref{sec:ab}, we conduct comprehensive ablation studies on both the semantic-aware token selection mechanism and the video parallel speculative decoding framework, initially employing video-SALMONN2+ as the backbone model. To rigorously validate the generalization and consistency of our proposed modules across diverse video-LLMs, we extend these ablation experiments to encompass three additional popular video-LLMs: Qwen2.5-VL, LLaVA-OneVision, and Qwen3-VL. This broader experimental scope enables us to establish the universal efficacy of our approach beyond a single model implementation.

  As illustrated in Figures \ref{figs:abla_qwen25vl}, \ref{figs:abla_llavaov}, and \ref{figs:abla_qwen3vl}, our ablation results consistently demonstrate that the removal of any individual component within the \method framework leads to measurable performance degradation across all evaluated backbone models, thereby validating the necessity and synergistic contribution of each module. When examining different pruning strategies in isolation, we observe that leveraging the spatial and temporal redundancy yields consistently effective results across all architectures.
   This empirical finding provides evidence that reliance solely on target model attention signals—without explicit vision-centric modeling—proves insufficient for effectively capturing the hierarchical and temporally-evolving semantic information essential for complex, long-form video understanding tasks. These results underscore the critical importance of our integrated spatiotemporal-aware pruning strategy in achieving both computational efficiency and semantic fidelity.

\section{More Results of Case Study} \label{app:case}

  In Section \ref{sec:prun}, we evaluate traditional vision-centric single-model pruning methods using video-SALMONN2+ as the backbone model on the VideoDetailedCaption dataset. Our preliminary findings reveal that pruning methods relying exclusively on signals from a single model consistently underperform compared to our proposed \method approach, which strategically incorporates position-debiased guidance derived from the target model. To comprehensively assess the generalization of these observations, we extend our comparative analysis to different video-LLMs.

  As illustrated in Figures \ref{figs:mat_qwen25}, \ref{figs:mat_llavaov}, and \ref{figs:mat_qwen3}, our extended experimental results corroborate the initial findings across all evaluated datasets. Specifically, single-model pruning strategies consistently exhibit inferior performance when contrasted with our proposed framework that integrates position-debiased guidance signals from the target model. We attribute this substantial and consistent performance gap to two fundamental factors. First, the draft model derives significant benefits from preserving the holistic temporal structure and distributional characteristics inherent in the original video sequence—properties that single-model attention-based pruning often fails to maintain due to its susceptibility to position bias. Second, the explicit incorporation of target model signals provides crucial cross-model guidance that enables superior alignment with the target model's internal representation space and decision-making process, thereby facilitating more effective speculative decoding.

  It is important to emphasize that single-model pruning techniques are not inherently incompatible with our \method. Consequently, investigating principled methods to synergistically combine these two paradigms—specifically, leveraging vision-centric redundancy signals to refine initial pruning decisions while simultaneously integrating target model guidance to ensure distributional alignment—represents a particularly promising avenue for future research. Such hybrid strategies hold substantial potential for constructing draft models that are simultaneously more computationally efficient and better aligned with target model semantics, thereby pushing the efficiency-accuracy frontier of video-LLM speculative decoding.

  \section{LLM Usage}
We used a large language model (LLM)–based writing assistant for grammar and wording improvements on draft text. The LLM did not generate research ideas, claims, proofs, algorithms, code, figures, or analyses, and it did not have access to any non-public data. During rationale generation, we use LLMs to transfer reasoning path into rationales.  All edits suggested by the LLM were manually reviewed and either accepted or rewritten by the authors, who take full responsibility for the final content. The LLM is not an author of this paper.
\end{document}